\documentclass[journal]{IEEEtran}
\usepackage{amsmath,amsfonts}
\usepackage{algorithmic}
\usepackage{algorithm}
\usepackage{array}
\usepackage{amssymb}  
\usepackage[caption=false,font=normalsize,labelfont=sf,textfont=sf]{subfig}
\usepackage{textcomp}
\usepackage{stfloats}
\usepackage{url}
\usepackage{verbatim}
\usepackage{graphicx}
\usepackage{cite}
\usepackage{cite}
\usepackage{booktabs}
\usepackage{multirow}
\usepackage[table,xcdraw]{xcolor}
\usepackage{hyperref}
\usepackage{cleveref}
\usepackage{marvosym}
\usepackage[normalem]{ulem}
\hypersetup{
    colorlinks=true,
    linkcolor=blue,
    citecolor=blue,
    urlcolor=blue
}

\begin{document}

\title{\textbf{RINO}: Accurate, Robust \textbf{R}adar-\textbf{I}nertial \textbf{O}dometry \\ with \textbf{N}on-Iterative Estimation*}

\author{Shuocheng Yang, Yueming Cao, Shengbo Eben Li, Jianqiang Wang\textsuperscript{\Letter} and Shaobing Xu\textsuperscript{\Letter}
\thanks{\Letter\ Corresponding author.}
}



\maketitle

\begin{abstract}
Odometry in adverse weather conditions, such as fog, rain, and snow, presents significant challenges, as traditional vision- and LiDAR-based methods often suffer from degraded performance. Radar-Inertial Odometry (RIO) has emerged as a promising solution due to its resilience in such environments. In this paper, we present RINO, a non-iterative RIO framework implemented in an adaptively loosely coupled manner. Building upon ORORA as the baseline for radar odometry, RINO introduces several key advancements, including improvements in keypoint extraction, motion distortion compensation, and pose estimation via an adaptive voting mechanism. This voting strategy facilitates efficient polynomial-time optimization while simultaneously quantifying the uncertainty in the radar module’s pose estimation. The estimated uncertainty is subsequently integrated into the maximum a posteriori (MAP) estimation within a Kalman filter framework. Unlike prior loosely coupled odometry systems, RINO not only retains the global and robust registration capabilities of the radar component but also dynamically accounts for the real-time operational state of each sensor during fusion. Experimental results conducted on publicly available datasets demonstrate that RINO reduces translation and rotation errors by 1.06\% and 0.09°/100m, respectively, when compared to the baseline method, thus significantly enhancing its accuracy. Furthermore, RINO achieves performance comparable to state-of-the-art methods. Our code is available at  \href{https://github.com/yangsc4063/rino}{https://github.com/yangsc4063/rino}.
\end{abstract}

\vspace{0.5em}

\def\abstractname{Note to Practitioners}
\begin{abstract}
Odometry for autonomous vehicles faces significant challenges under adverse conditions such as fog, rain, and snow. Commonly used sensors, including cameras and LiDAR, perform poorly in such environments, leading to failures in traditional odometry methods. Scanning millimeter-wave radar, with its strong penetration capabilities, offers a promising alternative by providing dense, wide-ranging, and robust perception under these conditions. However, radar data is often noisy and prone to ghosting artifacts, making processing challenging and hindering its widespread research and practical application. In this paper, we propose a Radar-Inertial Odometry (RINO) system that integrates pose estimation from scanning radar and IMU while incorporating uncertainty estimation within a non-iterative point cloud registration framework. Experimental validation on both datasets and real-world vehicle tests shows that RINO outperforms existing methods in terms of accuracy, efficiency, and robustness, providing a reliable solution for autonomous navigation in adverse weather conditions.
\end{abstract}


\begin{figure}
    \centering
    \includegraphics[width = \linewidth]{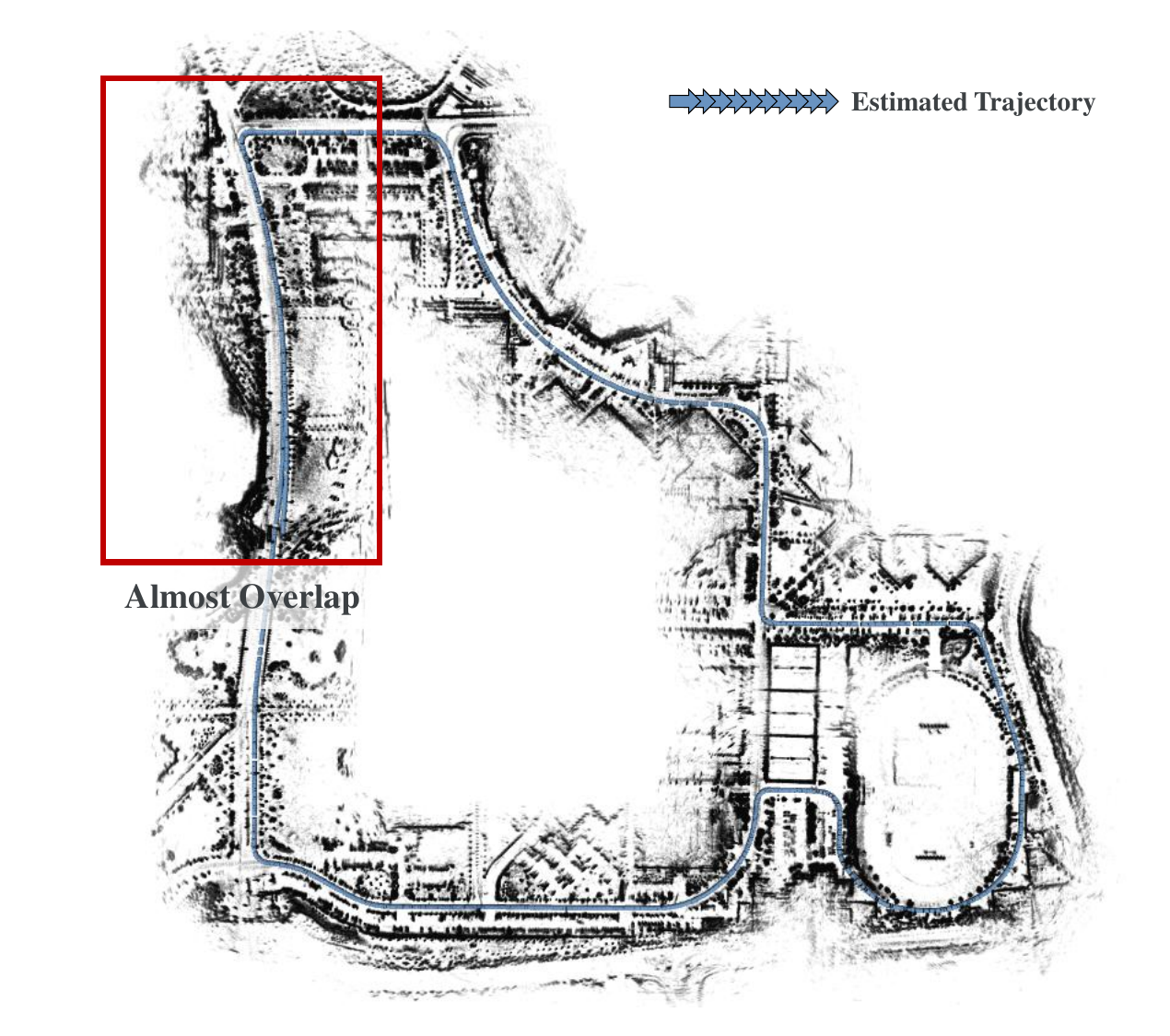}
    \caption{\textbf{Visualization of the trajectory} obtained by RINO on the \texttt{KAIST02} sequence of the Mulran dataset. The raw radar data and IMU data of the \texttt{KAIST02} sequence are given in the format of ROS messages using the File player provided by the Mulran dataset and can run robustly in real time after being processed by RINO. The trajectory of the second lap closely overlaps with that of the first lap, demonstrating the effectiveness of RINO.}
    \label{Fig:results}
\end{figure}

\section{INTRODUCTION} \label{Sec:Intro}

Autonomous driving systems rely on precise localization to ensure safe and efficient navigation. Over the past two decades, numerous odometry and SLAM methods have been developed based on cameras~\cite{qinVINSMonoRobustVersatile2018b, camposORBSLAM3AccurateOpenSource2021b, 10287884, 10488029, 10254455, 9926038} and LiDAR~\cite{zhangLOAMLidarOdometry2014, xuFASTLIOFastRobust2021, dellenbachCTICPRealtimeElastic2022, 9765591, 10721206}, all of which have demonstrated excellent performance. However, these methods are often limited by adverse weather conditions, such as fog, rain, or snow. Millimeter-wave radar, due to its longer operating wavelengths, can perform robustly in such conditions, making it a promising alternative. Nonetheless, radar data is inherently noisy and prone to significant distortion, posing substantial challenges for the development of effective odometry and SLAM algorithms.

Given these challenges, an increasing number of researchers are exploring solutions in radar odometry~\cite{abu-alrubRadarOdometryAutonomous2023}. Existing methods can be broadly classified into sparse and dense approaches. Dense methods directly process the tensor-format data acquired, whereas sparse methods~\cite{adolfssonCFEARRadarodometryConservative2021, adolfssonLidarLevelLocalizationRadar2022, zhangScanDenoisingNormal2023a, lubancoR3RobustRadon2024} extract keypoints from the radar tensor before performing registration. Sparse methods generally offer advantages in computational efficiency and accuracy. Some of these approaches draw inspiration from global pointcloud registration techniques in computer vision, reducing dependence on an accurate initial pose estimate and improving the stability of radar odometry. A notable example in this category is ORORA~\cite{limORORAOutlierRobustRadar2023a}. However, despite its robustness, ORORA's accuracy remains less competitive compared to other radar odometry solutions.

A natural approach to addressing these limitations is the integration of an IMU, which can compensate for motion distortions and enable a coupled odometry system to mitigate short-term pose drift. However, ORORA decouples rotation and translation when estimating pose transformations. To enhance robustness against outliers in non-convex optimization problems, it employs techniques such as graduated non-convexity (GNC). While these optimization strategies improve reliability, they also complicate the implementation of tightly coupled radar-inertial odometry (RIO). In contrast, loosely coupled approaches, although simpler, are often criticized for their inability to dynamically adapt to real-time sensor uncertainty.

To address these limitations, we propose a radar-inertial odometry (RIO) system, \textbf{RINO}, which employs a non-iterative pose estimation approach and quantifies the uncertainty of the pose transformation estimated by the radar branch. This enables the construction of a loosely coupled Kalman filter that dynamically adjusts fusion weights based on the real-time state of the sensors. RINO builds upon ORORA~\cite{limORORAOutlierRobustRadar2023a} as the baseline for the radar odometry branch, with several key improvements. We enhance keypoint extraction, compensate for motion distortions, and employ an adaptive voting strategy to solve the pose optimization problem, thereby avoiding iterative computations while obtaining a quantitative representation of pose estimation uncertainty. The estimated uncertainty of the radar module is then incorporated into the maximum a posteriori (MAP) estimation framework of the filter. Unlike previous methods, we introduce, for the first time in a loosely coupled system, a dynamic incorporation of each sensor’s real-time estimation quality rather than simply performing modular fusion of their outputs. Since the IMU is also resilient to adverse weather conditions, we argue that RINO can provide more robust pose estimation in complex environments, enhancing the reliability of radar odometry. The main contributions of this paper are as follows:

\begin{itemize}
    \item An adaptive loosely coupled RIO. Unlike conventional loosely coupled approaches that perform modular sensor fusion, RINO dynamically adjusts fusion weights based on the real-time estimation quality of each sensor, enhancing adaptability to varying sensor conditions.
    \item A non-iterative pose estimation with uncertainty quantification. RINO introduces a non-iterative pose estimation method that avoids costly iterative computations while simultaneously quantifying the uncertainty of the radar-based pose transformation.
    \item Extensive experiments on datasets demonstrate that RINO exhibits strong performance and adaptability across various scenarios and adverse weather conditions. Real-world vehicle tests further validate its practical applicability and reliability.
\end{itemize}

\section{RELATED WORKS} \label{Sec:Relat}

\subsection{Radar Odometry} \label{subSec:RO}

In the past decade, radar-based odometry has garnered significant attention from several research teams worldwide. Among the various radar technologies, 4D imaging radar and scanning radar have emerged as two promising candidates. Single-chip 4D imaging radars are compact, cost-effective, and capable of capturing elevation information of the scene. While researchers have explored odometry solutions~\cite{zhuang4DIRIOM4D2023a, zhang4DRadarSLAM4DImaging2023c, luEfficientDeepLearning4D2024, zhuo4DRVONetDeep4D2024} with these sensors, their sparse and noisy point clouds pose challenges for real-world deployment. In contrast, scanning radar provides higher azimuth resolution and denser raw data, making it more suitable for precise motion estimation or scene understanding. Therefore, this paper focuses on developing odometry methods based on scanning radar.

Scanning radar data is particularly intriguing, as it can be represented both as dense data, similar to images, and as sparse data by extracting keypoints akin to point clouds. Consequently, these research methods can be classified into two categories: dense methods and sparse methods.

Dense methods directly process raw or minimally preprocessed tensor-format data from scanning radar, originating from the direct method in visual odometry. Inspired by Fourier-Mellin Transformation (FMT), methods such as~\cite{parkPhaRaODirectRadar2020, barnesMaskingMovingLearning2020b, westonFastMbyMLeveragingTranslational2022a} have been developed. However, these methods require substantial computational resources and struggle with dynamic distortion and Doppler shift compensation.

Sparse methods generally involve two steps: first, extracting keypoints from raw images of scanning radar, and subsequently solving the pointcloud registration problem.

One typical sparse method originates from LiDAR odometry (LO), requiring iterative solving of the transformation and rotation between consecutive scans. Works such as~\cite{adolfssonCFEARRadarodometryConservative2021, adolfssonLidarLevelLocalizationRadar2022, kungNormalDistributionTransformBased2021, zhangScanDenoisingNormal2023a, alhashimiBFARImprovingRadar2024} have applied classic LO algorithms, such as point-to-line ICP or NDT, to scanning radar with satisfactory results. However, these iterative methods heavily rely on initial values and online optimizations, making them less robust.

The other type of sparse method stems from the indirect methods in visual odometry, typically utilizing feature descriptors to describe and match keypoints from two scans. However, descriptor-based matching inevitably introduces errors, leading to a high proportion of outliers among the keypoints. The presence of high noise and ”ghosting” artifacts in scanning radar exacerbates this issue. Therefore, the primary challenge of these methods is how to effectively identify and remove outliers. Hong et al.~\cite{hongRadarSLAMRadarBased2020b, hongRadarSLAMRobustSimultaneous2022} used distance consistency between pairwise points as a constraint to remove outliers and then used SVD decomposition to analytically solve for rotation. Barnes et al.~\cite{barnesRadarLearningPredict2020} employed a U-net~\cite{ronnebergerUNetConvolutionalNetworks2015} style CNN to simultaneously extract keypoints and generate per-point descriptors. While the keypoints matching after training is relatively accurate, its generalization performance is poor. Burnett et al.~\cite{burnettWeNeedCompensate2021} utilized the classic RANSAC method to remove outliers in keypoint matching and demonstrated the necessity of incorporating dynamic distortion compensation into scanning radar-based odometry.

Recently, researchers have recognized a major advantage of these methods: good transferability to pointcloud registration techniques active in the computer vision field. These techniques offer novel solutions for pointcloud registration with many outliers or low overlap. TEASER \cite{yangTEASERFastCertifiable2021} models the robust pointcloud registration problem using truncated least squares, decoupling the estimation of scaling, rotation, and translation. It applies Max Clique Inlier Selection (MCIS), Graduated Nonconvexity (GNC), and adaptive voting in a cascaded manner, significantly improving the accuracy, efficiency, and robustness of the solution. A recent representative study is ORORA~\cite{limORORAOutlierRobustRadar2023a}, which draws inspiration from TEASER and introduces a radar odometry pipeline based on the global pointcloud registration method. Additionally, it models the anisotropic uncertainty of keypoints extracted from scanning radar raw data in tensor format, leveraging this information as an initial estimate for rotation estimation and for anisotropic component-wise translation estimation (A-COTE). Compared to previous methods, ORORA does not rely on an accurate initial pose transformation and has demonstrated good accuracy and real-time performance on publicly available datasets. Given its robustness, ORORA represents a promising direction for advancing radar odometry.

Due to its robust pointcloud registration capabilities, we have chosen ORORA as the baseline for the radar branch. However, it still presents some areas for improvement. For example, (1) ORORA does not account for motion distortion, which is a critical factor influencing point cloud registration accuracy; (2) the rotation estimation method, GNC, still requires multiple iterations, resulting in unstable computation times and some dependence on initial values. In this paper, we investigate the impact of motion distortion on odometry results. Additionally, we propose an adaptive voting method for rotation estimation, which eliminates the need for iterative solving, ensures stable polynomial time complexity, and completely removes the reliance on good initial values.

\subsection{Radar-Inertial Odometry} \label{subSec:RIO}

Given our dissatisfaction with the odometry accuracy of ORORA, we consider incorporating an IMU to form a coupled odometry system. Several studies~\cite{liangScalableFrameworkRobust2020, dearaujoNovelMethodLand2023} have focused on ego-motion estimation by fusing traditional automotive radar and IMU using Kalman filtering (KF) and its various derivatives. Kalman filtering-based fusion can be broadly categorized into loosely coupled and tightly coupled methods. Tightly coupled methods integrate the scanning radar point cloud registration residuals into the observations; however, these approaches rely on the equivalence between the Invariant Extended Kalman Filter (IEKF) and the Gauss-Newton method, which conflicts with the non-iterative estimation employed in RINO.

Fortunately, we can estimate both the pose and its uncertainty simultaneously using the adaptive voting method. By feeding both the pose and uncertainty into the Maximum A Posteriori (MAP) estimation module of the KF, we have constructed an adaptive loosely coupled odometry system that dynamically adjusts the fusion weights based on the sensor states.
\begin{figure*}
    \centering
    \includegraphics[width = \linewidth]{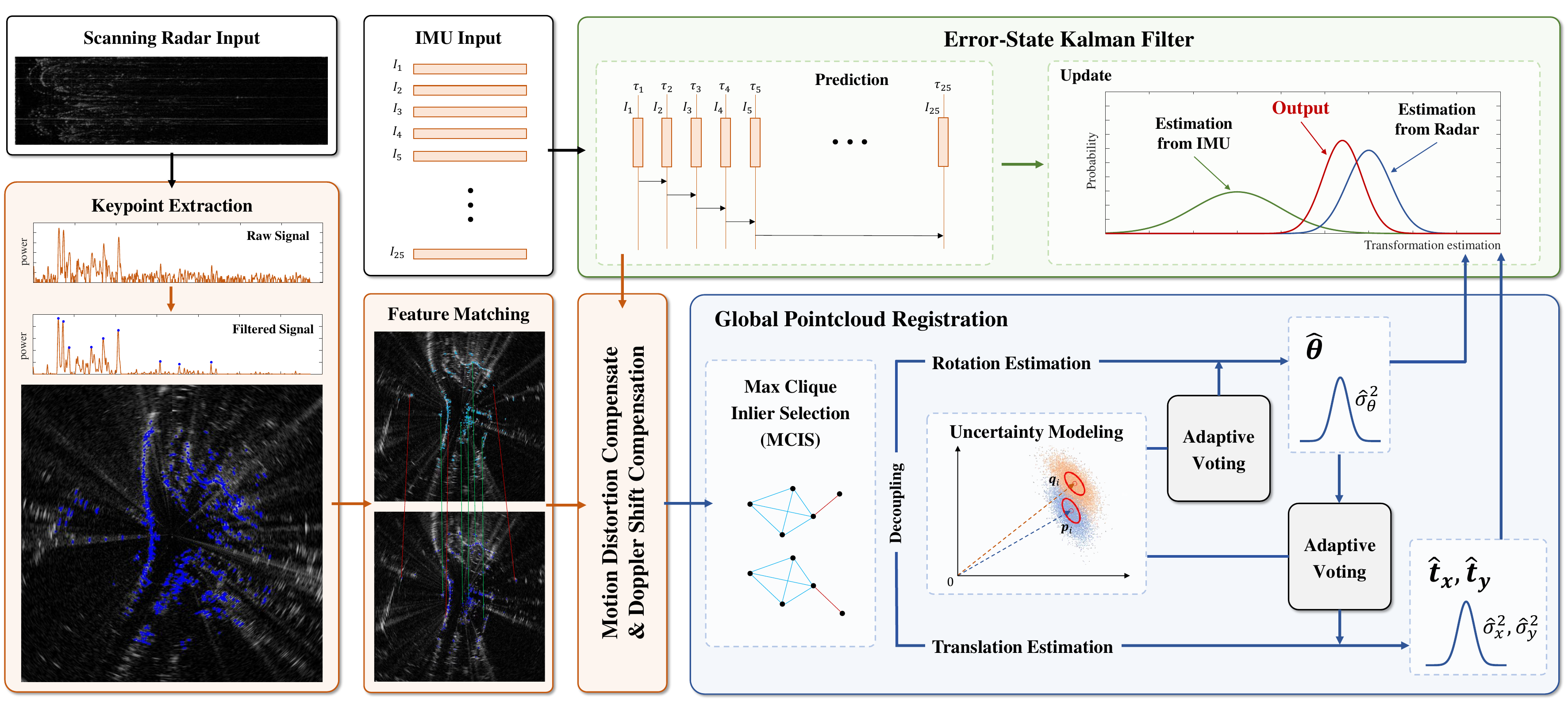}
    \caption{\textbf{Overview of RINO.} The IMU branch is responsible for constructing the motion equations and performing motion distortion compensation. The Radar branch extracts and matches keypoints from paired raw radar data. After MCIS, the Radar branch non-iteratively estimates the rotation and translation between two scans, along with their uncertainties. These estimates are then fused with the IMU's uncertainty input to the ESKF update step, leading to the optimal pose estimation through MAP.}
    \label{Fig:overview}
\end{figure*}

\section{METHODOLOGY} \label{Sec:Metho}

\subsection{System Overview} \label{subSec:Syste}

The proposed workflow called \textbf{RINO} is showed in Fig. \ref{Fig:overview}. It receives scanning radar (4Hz) and IMU data (100Hz) simultaneously. We aim to use observation of these two sensors to estimate the state of the vehicle. General state estimation can be formulated as a Maximum A Posteriori (MAP) problem. Assuming the motion process exhibits Markovian properties, this problem can be solved by filtering methods, which is advantageous due to its lightweight nature. In the filter, IMU measurements are used to construct the prediction equation as the prior estimate of the motion state, while the registration results from scanning radar branch serve as the observation for the pose. In the following, the two branches of RINO are introduced respectively.

\subsection{State Estimations} \label{subSec:IMU}
For the IMU, its state variables consist of:
\begin{equation}
    \boldsymbol{x}=[\boldsymbol{p},\boldsymbol{v},\boldsymbol{R},\boldsymbol{b}_g,\boldsymbol{b}_a,\boldsymbol{g}]^\top
\end{equation}
where $\boldsymbol{p}\in\mathbb{R}^3$ and $\boldsymbol{R}\in \mathrm{SO}(3)$ are the position and altitude of IMU, $\boldsymbol{v}\in\mathbb{R}^3$ is the velocity, $\boldsymbol{b}_g,\boldsymbol{b}_a\in\mathbb{R}^3$ is the IMU bias and $\boldsymbol{g}\in\mathbb{R}^3$ is gravity. These variables are relative to the world coordinate system. However, when constructing the motion equations with these state variables, linearizing the rotation variables becomes a challenging problem. Inspired by the FAST-LIO\cite{xuFASTLIOFastRobust2021}, we treat the original state variables as nominal state variables, where the difference between the nominal and true values represents the error state variables. Moreover, the handling of noise is incorporated into the error state variables. The kinematic equations for nominal state variables are
\begin{equation}
    \begin{aligned}
        \boldsymbol p(t+\Delta t)&=\boldsymbol p+\boldsymbol v\Delta t+\frac{1}{2}\big(\boldsymbol R(\boldsymbol a_I-\boldsymbol b_a)\big)\Delta t^2+\frac{1}{2}\boldsymbol g\Delta t^2,\\
        \boldsymbol v(t+\Delta t)&=\boldsymbol v+\boldsymbol {R}(\boldsymbol a_I-\boldsymbol b_a)\Delta t+\boldsymbol g\Delta t,\\
        \boldsymbol R(t+\Delta t)&=\boldsymbol R\cdot\mathrm{Exp}\big((\boldsymbol \omega_I-\boldsymbol b_g)\Delta t\big),\\
        \boldsymbol b_g(t+\Delta t)&=\boldsymbol b_g,\;
        \boldsymbol b_a(t+\Delta t)=\boldsymbol b_a,\\
        \boldsymbol g(t+\Delta t)&=\boldsymbol g.
    \end{aligned}
    \label{Eq:nominal}
\end{equation}
And error state variables in discrete time are
\begin{equation}
    \begin{aligned}
        \delta\boldsymbol p(t+\Delta t)&=\delta\boldsymbol p+\delta\boldsymbol v\Delta t,\\
        \delta\boldsymbol v(t+\Delta t)&=\delta\boldsymbol v\\
        &+\big(-\boldsymbol R(\boldsymbol a_I-\boldsymbol b_a)^\wedge\delta\boldsymbol \theta-\boldsymbol R\delta\boldsymbol b_a+\delta\boldsymbol g\big)\Delta t-\boldsymbol \eta_v,\\
        \delta\boldsymbol \theta(t+\Delta t)&=\mathrm{Exp}\big(-(\boldsymbol \omega_I-\boldsymbol b_g\Delta t\big)\delta\boldsymbol \theta-\delta\boldsymbol b_g\Delta t-\boldsymbol \eta_\theta,\\
        \delta\boldsymbol b_g(t+\Delta t)&=\delta\boldsymbol b_g+\boldsymbol \eta_g,\;
        \delta\boldsymbol b_a(t+\Delta t)=\delta\boldsymbol b_a+\boldsymbol \eta_a,\\
        \delta\boldsymbol g(t+\Delta t)&=\delta\boldsymbol g.
    \end{aligned}
    \label{Eq:error}
\end{equation}
where $\Delta t$ denotes the discrete time step. $\boldsymbol a_I$ and $\boldsymbol \omega_I$ are the measurements of IMU, while $\boldsymbol \eta_a$ and $\boldsymbol \eta_g$ are white noise of IMU measurements. $\boldsymbol \eta_v$ is the result of $\boldsymbol \eta_a$ acting on velocity, satisfying $\sigma(\boldsymbol \eta_v)=\Delta\sigma(\boldsymbol \eta_a)$, and likewise for $\boldsymbol \eta_\theta$.

Based on Eq.\eqref{Eq:nominal} and Eq.\eqref{Eq:error}, we perform the prediction process of the Error-State Kalman Filter (ESKF). Since the noise has been incorporated into the error states, the prediction of nominal state variables can be directly obtained from equation Eq.\eqref{Eq:nominal}. As for the error states,
\begin{equation}
    \begin{aligned}
        \delta\boldsymbol x_{\mathrm{pred}}&=\boldsymbol A\delta\boldsymbol x+\boldsymbol{\eta},\\
        \boldsymbol P_{\mathrm{pred}}&=\boldsymbol A\boldsymbol P\boldsymbol A^\top+\boldsymbol B.
    \end{aligned}
\end{equation}
where $\boldsymbol P$ is the estimate covariance matrix of error state variables $\delta\boldsymbol x$, $\boldsymbol{A}$ is the coefficient matrix obtained from expressing Eq.\eqref{Eq:error} in matrix multiplication form, and $\boldsymbol{B}$ is the noise matrix of $\delta\boldsymbol x$, where the covariance matrices of individual variables' noises are arranged diagonally.

Following the prediction step of the ESKF, the pose is computed at a high frequency of 100Hz, consistent with the IMU output frequency. These high-frequency yet coarse pose estimates can partially alleviate the motion distortions in single-frame data from the scanning radar. The specific methodology will be elaborated in Sec.\ref{subSec:Radar}.

In the update process of ESKF, the position and attitude output from RO can be considered as observations for the state variables in the RIO system. Let $\boldsymbol{p}_{\mathrm{radar}}\in\mathbb{R}^2$ and $\boldsymbol{R}_{\mathrm{radar}}\in\mathrm{SO}(2)$ represent the observations from RO at a certain moment. We can directly express them as observations to the error state variables.
\begin{equation}
    \delta\boldsymbol{z}=[\boldsymbol{p}_{\mathrm{radar}}-\boldsymbol{p},\ \mathrm{Log}(\boldsymbol R^\top\boldsymbol R_{\mathrm{radar}})]^\top,
\end{equation}
Consequently, the observation equation becomes remarkably simple:
\begin{equation}
    \begin{aligned}
        \boldsymbol p_{\mathrm{radar}}-\boldsymbol p&=\delta\boldsymbol{p}+\boldsymbol{\xi}_p,\\
        \mathrm{Log}(\boldsymbol R^\top\boldsymbol R_{\mathrm{radar}})&=\delta\boldsymbol{\theta}+\boldsymbol{\xi}_\theta,
    \end{aligned}
    \label{Eq:observation}
\end{equation}
where $\boldsymbol{\xi}_p$ and $\boldsymbol{\xi}_\theta$ are noise of observations from RO (In other words, the uncertainty of the scanning radar pose estimation). 
Similarly, write Eq.\eqref{Eq:observation} in its overall form:
\begin{equation}
    \delta\boldsymbol{z}=\boldsymbol{C}\delta\boldsymbol{x}+\boldsymbol{\xi},
\end{equation}
the observation matrix $\boldsymbol{C}$ can be easily provided as:
\begin{equation}
    \boldsymbol{C}=\begin{bmatrix}
        \boldsymbol{I}_3&\boldsymbol{0}&\boldsymbol{0}&\boldsymbol{0}&\boldsymbol{0}&\boldsymbol{0}\\
        \boldsymbol{0}&\boldsymbol{0}&\boldsymbol{I}_3&\boldsymbol{0}&\boldsymbol{0}&\boldsymbol{0}
    \end{bmatrix},
    \label{Eq:C}
\end{equation}
the noise matrix is $\boldsymbol{D}=\mathrm{diag}\big(\mathrm{Cov}(\boldsymbol{\xi}_p),\mathrm{Cov}(\boldsymbol{\xi}_\theta)\big)$. With these steps, the update process of ESKF can be executed as follows:
\begin{equation}
    \begin{aligned}
        \boldsymbol K&=\boldsymbol P_{\mathrm{pred}}\boldsymbol H^\top(\boldsymbol{C}\boldsymbol P_{\mathrm{pred}}\boldsymbol{C}^\top+\boldsymbol{D})^{-1},\\
        \delta\hat{\boldsymbol{x}}&=\delta\boldsymbol{x}_{\mathrm{pred}}+\boldsymbol{K}(\delta\boldsymbol{z}-\boldsymbol{C}\delta\boldsymbol{x}_{\mathrm{pred}}),\\
        \boldsymbol{P}&=(\boldsymbol{I}-\boldsymbol{K}\boldsymbol{C})\boldsymbol{P}_{\mathrm{pred}}.
    \end{aligned}
\end{equation}
where $c(\cdot)$ represents the observation function.

After prediction and update steps, it's necessary to incorporate the error states into the nominal states, i.e., to reset the ESKF (setting $\delta x=0$). It's important to note that resetting the error state variables will result in a change in the zero point of the rotation variable's tangent space. Therefore, the covariance matrix should be adjusted accordingly
\begin{equation}
    \begin{aligned}
        \boldsymbol J&=\mathrm{diag}(\boldsymbol{I}_{3},\boldsymbol{I}_{3},\boldsymbol{I}_{3}-\frac{1}{2}\delta\boldsymbol{\theta}^\wedge,\boldsymbol{I}_{3},\boldsymbol{I}_{3},\boldsymbol{I}_{3}),\\
        \boldsymbol{P}_{\mathrm{reset}}&=\boldsymbol{J}\boldsymbol{P}\boldsymbol{J}^\top
    \end{aligned}
\end{equation}
where $\delta \boldsymbol{\theta}$ represents the estimated error state variables before resetting, and $(\cdot)^\wedge$ denotes expressing the vector as a skew-symmetric matrix.

\subsection{Scanning Radar Odometry} \label{subSec:Radar}

\begin{figure}
    \centering
    \includegraphics[width = \linewidth]{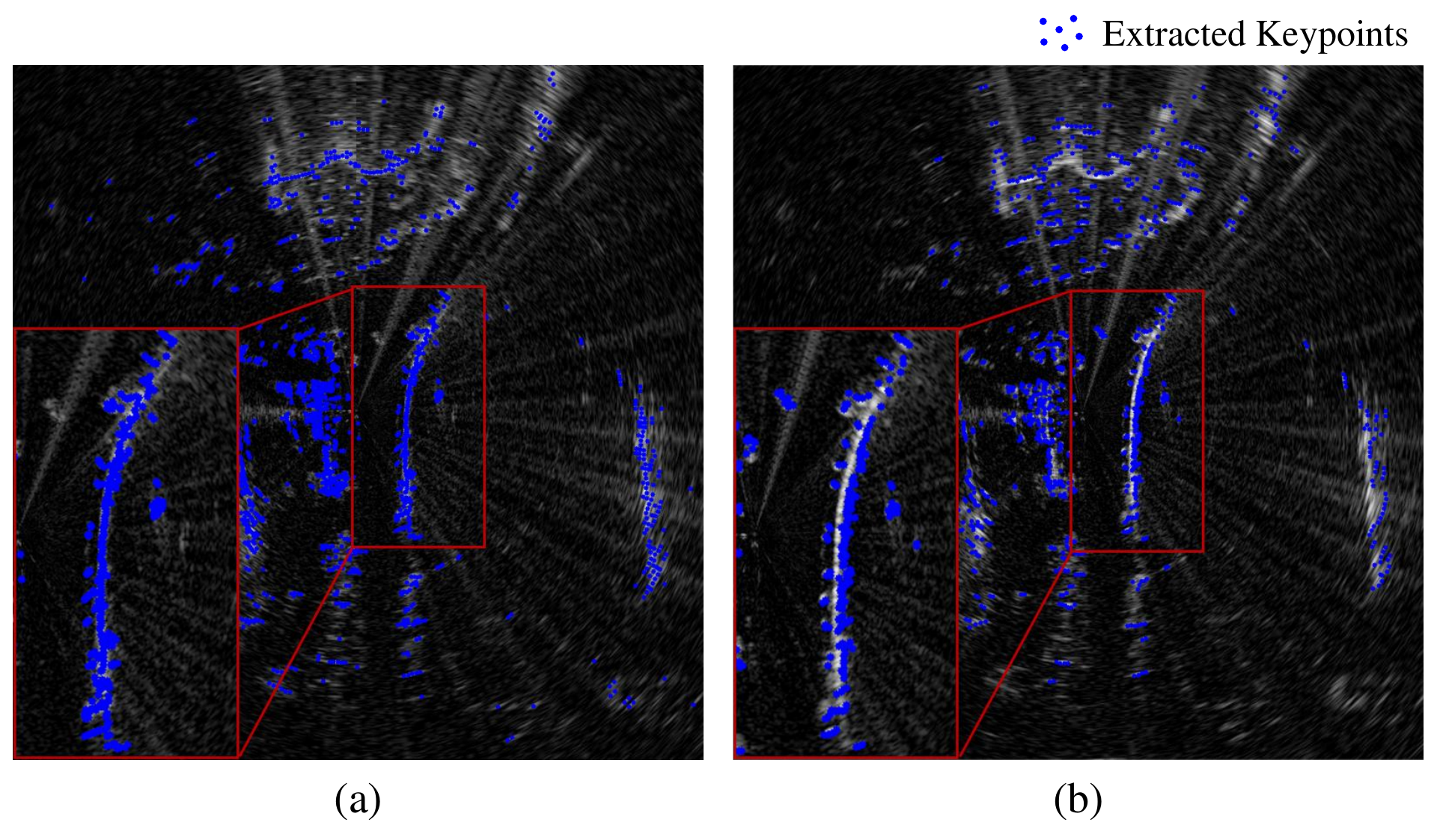}
    \vspace{-6mm}
    \caption{\textbf{Visualization of the keypoint extraction results.} The blue points represent the extracted keypoints, with the background displaying the Cartesian radar tensor. The grayscale values in the radar image correspond to the power of the radar echoes. (a) illustrates the results of the improved cen2018 method; (b) shows the results of the cen2019 method. It is clearly observable that the keypoints extracted by the improved cen2018 are better aligned with the brighter regions in the radar image.}
    \label{Fig:keypoint_extraction}
\end{figure}

\begin{figure}
    \centering
    \includegraphics[width = \linewidth]{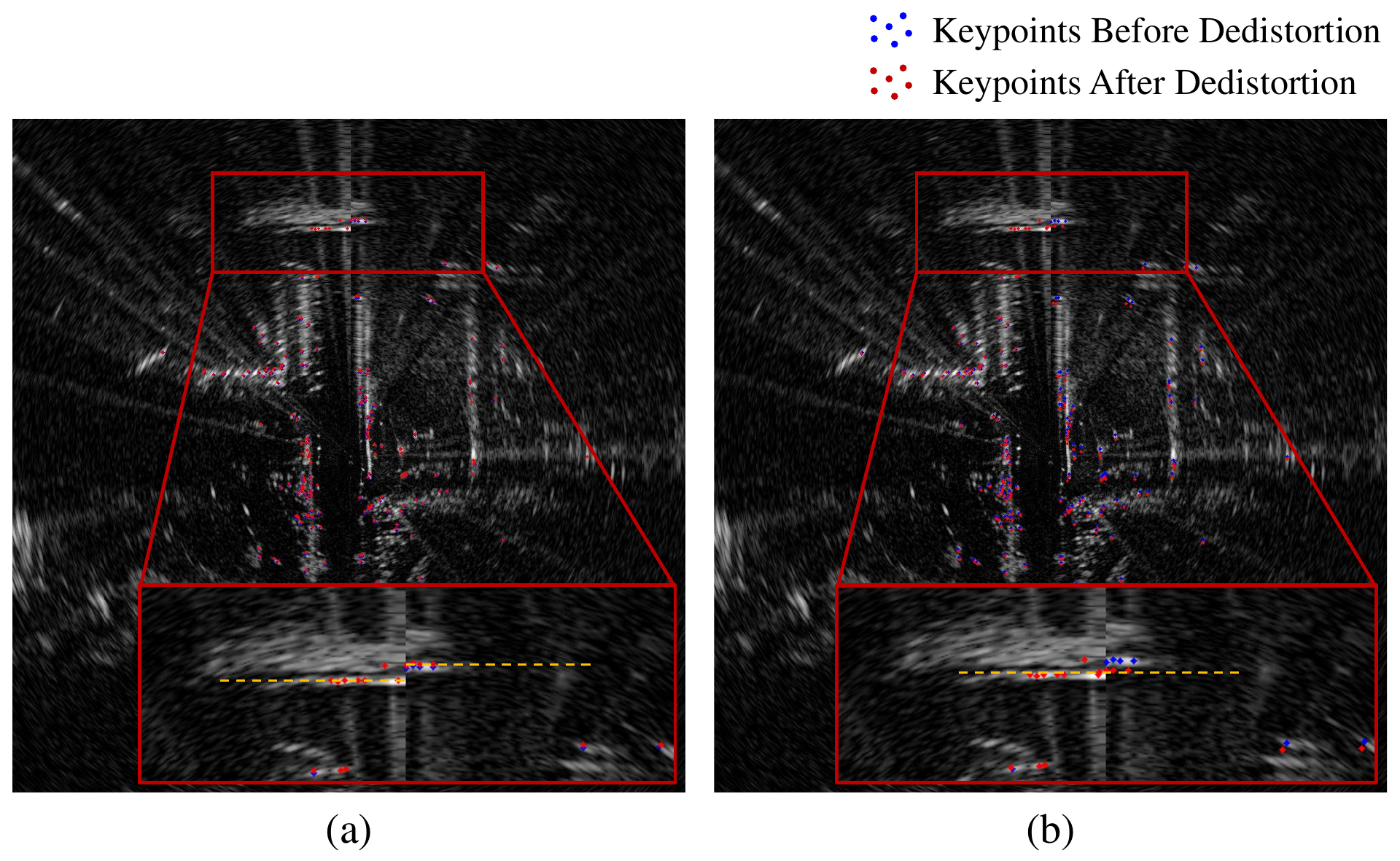}
    \vspace{-6mm}
    \caption{\textbf{Visualization of motion distortion compensation results.} The blue points represent the extracted keypoints, while the red points indicate the keypoints after motion distortion compensation. (a) shows the result with only Doppler shift compensation; (b) shows the result with both Doppler shift and motion distortion compensation. Note the zoomed-in region on the radar image, where it can be inferred that the red points correspond to keypoints extracted from a wall-like object. It is observable that before distortion compensation, the red points were split into two segments, while after compensation, the red points were restored to nearly a straight line.}
    \label{Fig:motion_distortion}
\end{figure}

RO of the system follows a similar workflow to ORORA \cite{limORORAOutlierRobustRadar2023a} which serves as the baseline. Firstly, keypoints are extracted from the raw intensity-distance-angle data of the scanning radar. Then, ORB\cite{rubleeORBEfficientAlternative2011} are used to generate descriptors for each point, followed by brute-force matching to establish correspondences set $\lbrace(\boldsymbol{p}_i,\boldsymbol{q}_i)\rbrace_{i=1}^N$ between keypoints from the previous scan $\lbrace\boldsymbol p_i\rbrace_{i=1}^N$ and current scan $\lbrace\boldsymbol q_i\rbrace_{i=1}^N$. Subsequently, compensation is applied to these correspondences for motion distortions and Doppler shifts. The Doppler compensation is applied to the radial distances of the keypoints, following \cite{limORORAOutlierRobustRadar2023a}. Finally, the compensated point correspondences are used for pointcloud registration, and the transformations from scan to scan are accumulated. We decouple the estimation of rotation and translation and employ an accelerated adaptive voting method, which enables fast computation of rotation and translation without the need for iteration. In the following, we focus on explaining the parts that differ from the baseline, while the remaining sections are detailed in\cite{limORORAOutlierRobustRadar2023a}.

The quality of keypoints extracted from radar directly determines the clarity of the resulting pointcloud map. Upon evaluating cen2018\cite{cenPreciseEgoMotionEstimation2018} and cen2019\cite{cenRadaronlyEgomotionEstimation2019}, we found that the keypoint positions obtained from cen2018 were significantly more accurate. Consequently, cen2018 was selected for our application, and we impose a limit on the maximum distance for it, because points too far away exhibit high uncertainty, negatively affecting pointcloud registration. Moreover, a high-pass filtering to the raw data was applied to augment the distinctiveness of signal peaks, thereby refining the keypoint extraction process.

As mentioned earlier, we use high-frequency pose estimates from the IMU to compensate for motion distortions and Doppler frequency shifts. Let $t_{s}$ denote the time for a single scan of the scanning radar. During $0\sim t_{s}$, 25 pose estimates $\boldsymbol{T}_1,\boldsymbol{T}_2,...,\boldsymbol{T}_{\mathrm{end}}$ from the IMU are received, each timestamped as $\tau_1,\tau_2,...,\boldsymbol{\tau}_{\mathrm{end}}$. By interpolating $\boldsymbol{T}_1,\boldsymbol{T}_2,...,\boldsymbol{T}_{\mathrm{end}}$, we can query the radar pose at any time within $0\sim t_{s}$. Therefore, we can transform the coordinates of any keypoint to $\tau_{\mathrm{end}}$ to achieve motion compensation.
\begin{equation}
    \begin{aligned}
        \boldsymbol{p}'=\boldsymbol{T}_{IR}^{-1}\boldsymbol{T}_{\mathrm{end}}^{-1}\boldsymbol{T}_\tau\boldsymbol{T}_{IR}\boldsymbol{p}_\tau
    \end{aligned}
\end{equation}
Here, $\boldsymbol{p}_\tau$ is the coordinates of the keypoint at $\tau$, $\boldsymbol{T}_\tau$ the interpolated IMU pose at $\tau$, and $\boldsymbol{T}_{IR}$ represents the extrinsic parameters between the IMU and the radar.

\subsection{Pointcloud Registration With Non-iterative Solving}

In Sec.\ref{subSec:Radar}, it is mentioned that the correspondences within $\lbrace(\boldsymbol{p}_i,\boldsymbol{q}_i)\rbrace_{i=1}^N$ are used to perform pointcloud registration to achieve scan-to-scan odometry. We focus on explaining this part in detail below.

We describe the rubust registration problem for scanning radar as follows:
\begin{equation}
    \begin{aligned}
        \boldsymbol{q}_i=\boldsymbol{R}^\circ\boldsymbol{p}_i+\boldsymbol{t}^\circ+\boldsymbol{o}_i+\boldsymbol{\varepsilon}_i
    \end{aligned}
\end{equation}
where $\boldsymbol{R}^\circ$ and $\boldsymbol{t}^\circ$ respectively denote the true values of rotation and translation. $\boldsymbol{o}_i$ indicates whether the correspondence for the i-th point is correct; if correct, $\boldsymbol{o}_i=0$, otherwise $\boldsymbol{o}_i$ can take any value. $\boldsymbol{\varepsilon}_i$ models the measurement noise.

Due to the presence of excessive noise and "ghosting" in the raw data from the scanning radar, the correspondence set $\lbrace(\boldsymbol{p}_i,\boldsymbol{q}_i)\rbrace_{i=1}^N$ contains a multitude of erroneous matches, as illustrated in the schematic diagram found in \cite{limORORAOutlierRobustRadar2023a}. Therefore, we adopt truncated least squares (TLS) estimation to establish an optimization model for rubust pointcloud registration, aiming to reduce sensitivity to outliers in the results as much as possible.
\begin{equation}
    \begin{aligned}
        \min_{\boldsymbol{R}\in \mathrm{SO}(2), \boldsymbol{t}\in\mathbb{R}^2}\sum_{i=1}^N\min\bigg\{\frac{1}{\beta_i^2}\|\boldsymbol{q}_i-\boldsymbol{R}\boldsymbol{p}_i-\boldsymbol{t}\|^2,\bar{c}^2\bigg\},
    \end{aligned}
    \label{Eq:tls}
\end{equation}
It is worth noting that $\beta_i$ in the Eq.\eqref{Eq:tls} represents the primary variance of the noise $\boldsymbol{\varepsilon}_i$ (the upper bound of the variance after projecting the two-dimensional Gaussian distribution onto each direction). This implies that the model may appear rather coarse when describing keypoints from scanning radar with significant anisotropic uncertainty.

To better capture the uncertainty of keypoints and reduce problem complexity, we decouple the rotation and translation. In detail, we subtract pairs of points within the same scan of pointcloud to eliminate the translation transformation, resulting in translation-invariant measurements (TIMs).
\begin{equation}
    \begin{aligned}
        \boldsymbol q_j-\boldsymbol q_i&=\boldsymbol R^\circ(\boldsymbol p_j-\boldsymbol p_i)+(\boldsymbol o_j-\boldsymbol o_i)+(\boldsymbol \varepsilon_j-\boldsymbol \varepsilon_i),\\
        \boldsymbol{q}_{ij}&=\boldsymbol{R}^\circ\boldsymbol{p}_{ij}+\boldsymbol{o}_{ij}+\boldsymbol{\varepsilon}_{ij},i,j=1,...,N,i\neq j
    \end{aligned}
    \label{Eq:TIMs}
\end{equation}
Then, we calculate the magnitude of the translation-invariant measurements to obtain translation-rotation invariant measurements (TRIMs).
\begin{equation}
    \begin{aligned}
        \|\boldsymbol q_{ij}\|&=\|\boldsymbol R^\circ\boldsymbol p_{ij}+\boldsymbol o_{ij}+\boldsymbol \varepsilon_{ij}\|,\\
        \|\boldsymbol q_{ij}\|&=\|\boldsymbol p_{ij}\|+\widetilde{o}_{ij}+\widetilde{\varepsilon}_{ij},i,j=1,...,N,i\neq j
    \end{aligned}
\end{equation}
where $\widetilde{o}_{ij}\leq\|\boldsymbol o_{ij}\|$, and $\widetilde{\varepsilon}_{ij}\leq\|\boldsymbol \varepsilon_{ij}\|$. 

We first use TRIMs $\lbrace(\|\boldsymbol{p}_k\|,\|\boldsymbol{q}_k\|)\rbrace_{k=1}^{N(N-1)/2}$ for preliminary inlier selection, where we only need to check if $\big\|\|\boldsymbol q_k\| - \|\boldsymbol p_k\|\big\|$ is less than a predefined threshold. Like \cite{hongRadarSLAMRadarBased2020b}, we also employ maximum clique search to address this issue. For detailed descriptions, please refer to their works.

Next, we can use TIMs $\lbrace(\boldsymbol{p}_k,\boldsymbol{q}_k)\rbrace_{k=1}^{K},\ K\leq N(N-1)/2$ to estimate the rotation transformation, and then use $\hat{\boldsymbol{R}}$ to estimate the two components of the translation transformation separately.
\begin{equation}
    \begin{aligned}
        \hat{\boldsymbol{R}}&=\mathop{\arg\min}\limits_{\boldsymbol{R}\in SO(2)}\sum_{k=1}^K\min\bigg\{\frac{1}{\sigma_{R,k}^2}\|\boldsymbol q_k-\boldsymbol p_k\|^2,\bar{c}^2\bigg\},
    \end{aligned}
    \label{Eq:rot_est}
\end{equation}
\begin{equation}
    \begin{aligned}
        \hat{t}_j&=\mathop{\arg\min}\limits_{t_j\in\mathbb{R}}\sum_{i=1}^N\min\bigg\{\frac{1}{\sigma_{j,i}^2}\big(t_j-[\boldsymbol q_i-\hat{\boldsymbol R}\boldsymbol p_i]_j\big)^2,\bar{c}^2\bigg\},j=1,2
    \end{aligned}
    \label{Eq:trans_est}
\end{equation}

ORORA adopts the Graduated Non-Convexity (GNC) approach for rotation estimation, an efficient technique in addressing non-convex optimization challenges. However, GNC’s iterative nature can result in unstable computation times and a potential to converge to local minima, which may introduce discontinuities in trajectory estimation. To mitigate these issues, we express the optimization problems for rotation and translation estimation in a unified form and implement a non-iterative solution using an adaptive voting method. Furthermore, we obtained the variances of the rotation and translation estimates to serve as inputs for the update process of ESKF.

It can be observed that a two-dimensional rotation can be expressed as a scalar (rotation angle). Therefore, Eq.\eqref{Eq:TIMs} and Eq.\eqref{Eq:rot_est} can be rewritten as:
\begin{equation}
    \begin{aligned}
        \mathrm{Ang}(\boldsymbol{q}_k)&=\mathrm{Ang}(\boldsymbol{p}_k)+\theta^\circ+\mathrm{Ang}(\boldsymbol{o}_k)+\mathrm{Ang}(\boldsymbol{\varepsilon}_k),
    \end{aligned}
\end{equation}
\begin{equation}
    \begin{aligned}
        \hat{\theta}&=\mathop{\arg\min}\limits_{\theta\in \mathbb{R}}\sum_{k=1}^K\min\bigg\{\frac{1}{\sigma_{\theta,k}^2}\big[\theta-\big(\mathrm{Ang}(\boldsymbol q_k)-\mathrm{Ang}(\boldsymbol p_k)\big)\big]^2,\bar{c}^2\bigg\}
    \end{aligned}
    \label{Eq:theta_est}
\end{equation}
Where $\mathrm{Ang}(\cdot)$ represents the function to compute the argument (angle) of a vector.

It is revealed that the objective functions in Eq.\eqref{Eq:theta_est} and Eq.\eqref{Eq:trans_est} actually have the same form: the sum of truncated quadratic functions with scalar variables. For such optimization problems, two key issues need to be considered. First, how to estimate the variance of decision variables. Second, how to solve the optimization problem itself.

\begin{figure}
    \begin{minipage}[t]{\linewidth}
        \centering
        \includegraphics[width=\textwidth]{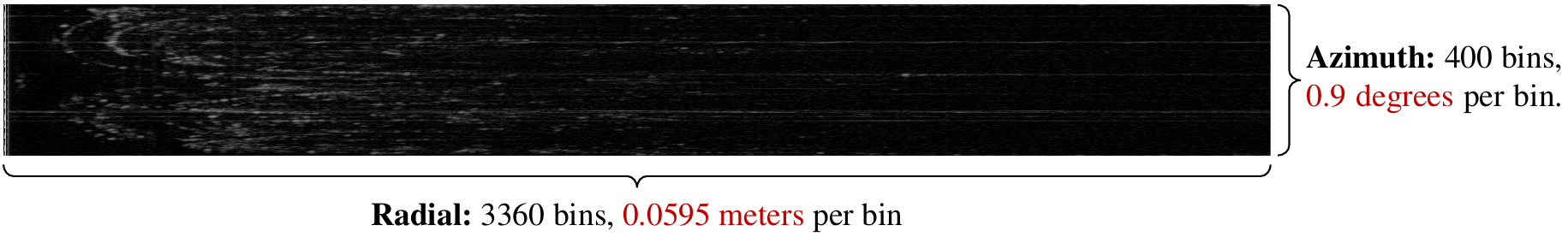}
        \centerline{(a)}
    \end{minipage}
    \begin{minipage}[t]{\linewidth}
        \centering
        \includegraphics[width=\textwidth]{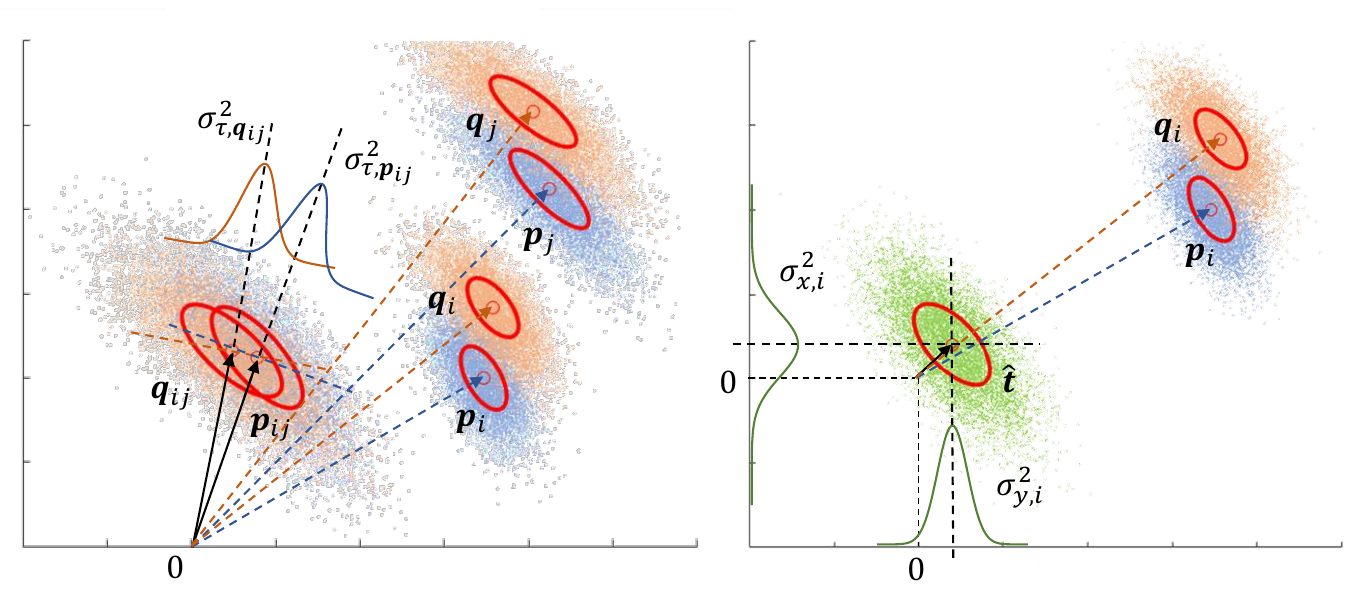}
        \centerline{(b)}
    \end{minipage}
    \caption{\textbf{Uncertainty model of rotation and translation estimates.} (a) The raw data of scanning radar clearly exhibits significantly higher resolution in the radial direction compared to the angular direction. This results in severe anisotropic uncertainty in the keypoints extracted from it. (b) The variances of the rotation estimates (left) and translation estimates (right) are calculated from the uncertainty model of the keypoints.}
    \label{Fig:uncertainty}
\end{figure}

Considering the first issue, the variances of rotation and translation originate from the uncertainty of keypoints.
\begin{equation}
    \begin{aligned}
        \theta_k=\mathrm{Ang}(\boldsymbol{q}_k)-\mathrm{Ang}(\boldsymbol{p}_k),\ k=1,...,K,
    \end{aligned}
    \label{Eq:rot_k}
\end{equation}
\begin{equation}
    \begin{aligned}
        \boldsymbol{t}_i=\boldsymbol{q}_i-\hat{\boldsymbol{R}}\boldsymbol{p}_i,\ i=1,...,N,
    \end{aligned}
    \label{Eq:trans_i}
\end{equation}
Therefore, our discussion should start with modeling the uncertainty of keypoints. ORORA’s modeling of the anisotropic uncertainty of keypoints is recognized for its effectiveness and is therefore adopted. We assume that the uncertainty of keypoint $\boldsymbol{p}_i(\rho_i,\phi_i)$ follows independent normal distributions in the range and azimuth $\rho_i\sim\mathcal{N}(\rho_i^\circ, \sigma_{\rho}^2), \phi_i\sim\mathcal{N}(\phi_i^\circ, \sigma_{\phi}^2)$, where $\sigma_{\rho}^2$ and $\sigma_{\phi}^2$ can be estimated from the raw polar coordinate data of the scanning radar (Fig.\ref{Fig:uncertainty}(a)). Following the derivation in \cite{limORORAOutlierRobustRadar2023a}, keypoint $\boldsymbol{p}_i$ is modeled as a normal distribution $\boldsymbol{p}_i\sim\mathcal{N}(\boldsymbol{p}_i, \boldsymbol{C}_{\boldsymbol{p}_i})$, where the value of $\boldsymbol{C}_{\boldsymbol{p}_i}$ can also be found in \cite{limORORAOutlierRobustRadar2023a}.

For rotation, the covariance of TIMs' distribution can be computed from $\boldsymbol{C}_{\boldsymbol{p}_{ij}}=\boldsymbol{C}_{\boldsymbol{p}_{i}}+\boldsymbol{C}_{\boldsymbol{p}_{j}}$, and then projected onto the tangential direction, as shown in the left plot of Fig.\ref{Fig:uncertainty}(b).
\begin{equation}
    \begin{aligned}
        \boldsymbol{C}_{\boldsymbol{p}_{ij}}'=\boldsymbol{W}\boldsymbol{C}_{\boldsymbol{p}_{ij}}\boldsymbol{W}^\top
    \end{aligned}
\end{equation}
Here, $\boldsymbol{W}$ is the matrix that transforms the covariance from the original coordinate to the one with $\boldsymbol{p}_{ij}$ as the axis. Then, we take the element (1,1) of $\boldsymbol{C}_{ij}'$, denoted as $\sigma_{\tau,\boldsymbol{p}_{ij}}^2$, and divide it by the magnitude of $\boldsymbol{p}_{ij}$ to approximate the variance of the distribution of $\mathrm{Ang}(\boldsymbol{p}_{ij})$. Then, according to Eq.\eqref{Eq:rot_k}, the variance of the rotation can be found:
\begin{equation}
    \begin{aligned}
        \sigma_{\theta,k}^2=\frac{\sigma_{\tau,\boldsymbol{p}_{ij}}^2}{\|\boldsymbol{p}_{ij}\|^2}+\frac{\sigma_{\tau,\boldsymbol{q}_{ij}}^2}{\|\boldsymbol{q}_{ij}\|^2}
    \end{aligned}
\end{equation}

For translation, we need to utilize the previously obtained rotation $\hat{\boldsymbol{R}}$. From Eq.\eqref{Eq:trans_i}, the covariance of $\boldsymbol{t}_i$ can be expressed as $\boldsymbol{C}_{\boldsymbol{t}_i}=\boldsymbol{C}_{\boldsymbol{q}_i}+\hat{\boldsymbol{R}}\boldsymbol{C}_{\boldsymbol{p}_i}\hat{\boldsymbol{R}}^\top$. Next, as shown in the right plot of Fig.\ref{Fig:uncertainty}(b), we project the distribution of translation onto the $x$ and $y$ directions. In other words, we take the (1,1) and (2,2) elements of $\boldsymbol{C}_{\boldsymbol{t}_i}$, which respectively correspond to the variance of the first and second components of translation.

\begin{figure}
    \centering
    \includegraphics[width = \linewidth]{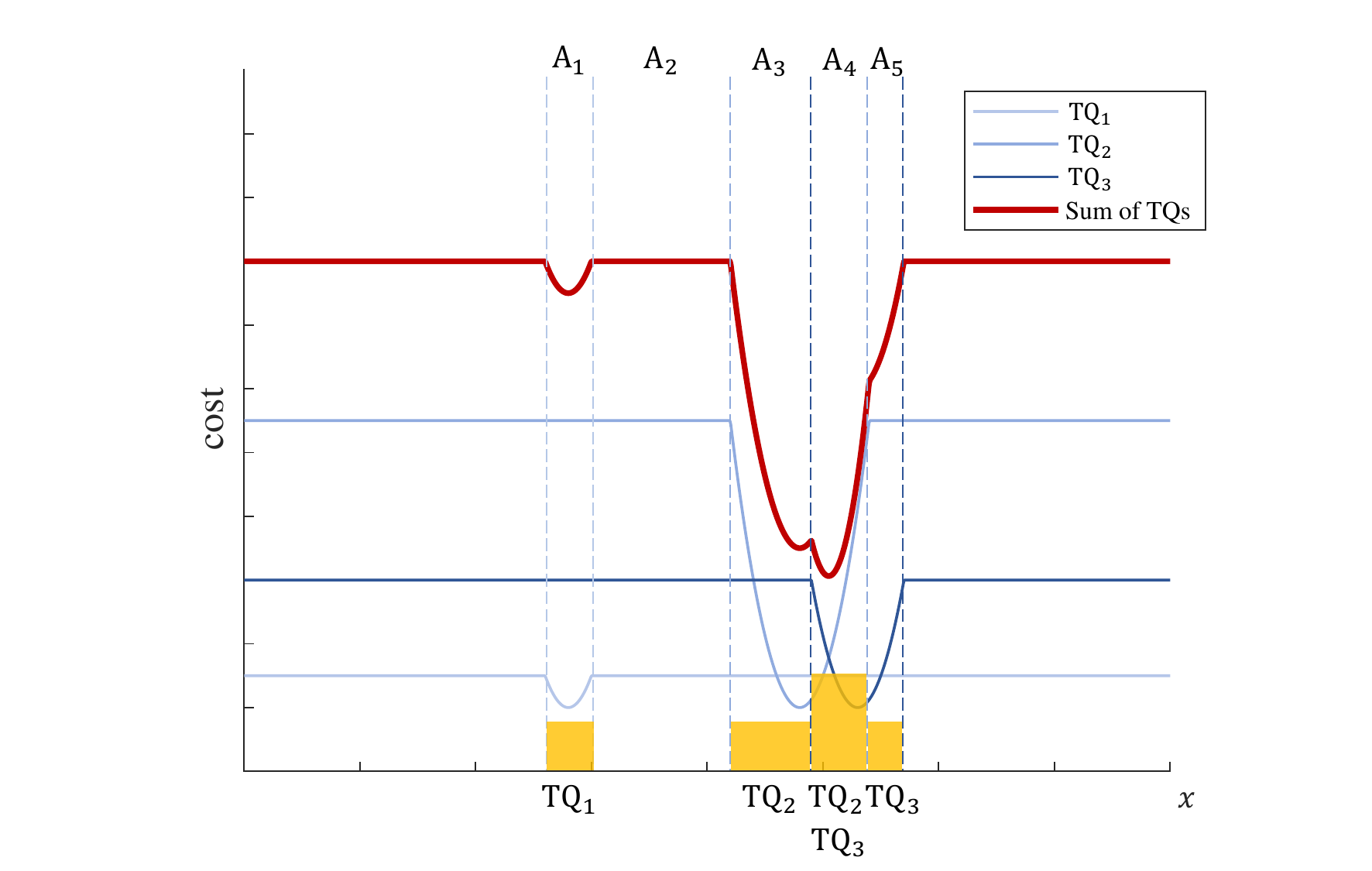}
    \caption{\textbf{Illustration of the adaptive voting method.} The sum of three truncated quadratic functions. The truncation points of the same function are denoted by vertical lines of the same color. The entire domain is partitioned into a disjoint union of seven sets, among which five are non-trivial ($A_1,...,A_5$). Enumerating the local minima of the convex functions over $A_1,...,A_5$, it can be inferred that the local minimum on $A_4$ is the global optimal solution.}
    \label{Fig:ad_voting}
\end{figure}

At last, we solve the optimization where the objective function is the sum of truncated quadratic functions. This is the key to solving the pointcloud registration with non-iterative method. Express equations Eq.\eqref{Eq:trans_est} and Eq.\eqref{Eq:theta_est} in a unified form:
\begin{equation}
    \begin{aligned}
        \hat{x}&=\mathop{\arg\min}\limits_{x\in \mathbb{R}}\sum_{m=1}^M\min\bigg\{\frac{1}{\sigma_{x,m}^2}(x-x_m)^2,\bar{c}^2\bigg\},m=1,...,M
    \end{aligned}
\end{equation}

From Fig.\ref{Fig:ad_voting}, we observe that although the objective function is globally highly non-convex, it is effectively piecewise convex. By partitioning the domain according to truncation points $\{x_m-\sigma_{x,m}^2\bar{c}^2,x_m+\sigma_{x,m}^2\bar{c}^2\}_{m=1}^M$, the objective function can be constrained to be convex (even quadratic) on each segment. Enumerating all segments, minimizing the objective function on each segment, and selecting the minimum value among all local minima enable the determination of the global minimum.

Let $(x-x_m)^2/\sigma_{x,m}^2$ be denoted as $f_m(x)$, and let the aforementioned segment be denoted as $\{A_n\}_{n=1}^{2M-1}$. Our objective is to compute the index set $I_n=\{m:f_m(x)\leq\bar{c}^2\}$ corresponding to each $A_n$. We have
\begin{equation}
    \begin{aligned}
        \min_{x\in \mathbb{R}}\sum_{m=1}^M\min\big\{f_m(x),\bar{c}^2\big\}=\min_n\min_{x\in \mathbb{R}}\sum_{r\in I_n}^Mf_r(x).
    \end{aligned}
\end{equation}
By enumerating to obtain the optimal segments as $\hat{n}$, the optimal estimator $\hat{x}$ we seek is
\begin{equation}
    \begin{aligned}
        \hat{x}=\bigg(\sum_{r\in I_{\hat{n}}}\frac{1}{\sigma_{x,r}^2}\bigg)^{-1}\sum_{r\in I_{\hat{n}}}\frac{x_r}{\sigma_{x,r}^2}.
    \end{aligned}
    \label{Eq:hat_x}
\end{equation}

\subsection{Pose uncertainty estimation}

We found that RINO may encounter some extreme situations during operation. For example, the extracted feature points might be too sparse, or there might be too many dynamic objects between two frames, resulting in a small number of inlier points remaining after MCIS, as shown in Fig.\ref{Fig:overview}. These issues could lead to a decrease in the confidence of the current scanning radar branch pose estimation. In such cases, we hope that RINO will be more inclined to trust the pose estimation from the IMU.

Naturally, we can estimate the variance of $\hat{x}$ from Eq.\eqref{Eq:hat_x}.
\begin{equation}
    \begin{aligned}
        \sigma_{\hat{x}}^2=\bigg(\sum_{r\in I_{\hat{n}}}\frac{1}{\sigma_{x,r}^2}\bigg)^{-1}.
    \end{aligned}
    \label{Eq:uncertainty}
\end{equation}

In summary, the aforementioned adaptive voting is used to estimate rotation angle and the two components of translation separately, along with their uncertainty. Subsequently, these three estimators and their uncertainty can be input as observation into the update process of the Kalman filter in Sec.\ref{subSec:IMU}, thereby completing the adaptive, loosely coupled state estimation between the scanning radar and the IMU.
\begin{figure*}
    \begin{minipage}[t]{0.33\linewidth}
        \centering
        \includegraphics[width=\textwidth]{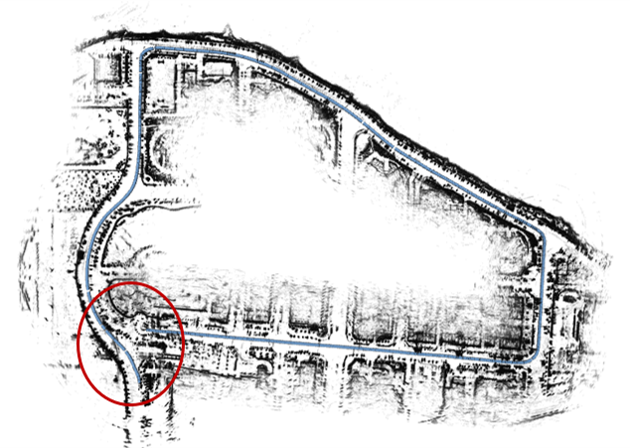}
        \centerline{(a)}
    \end{minipage}
    \begin{minipage}[t]{0.33\linewidth}
        \centering
        \includegraphics[width=\textwidth]{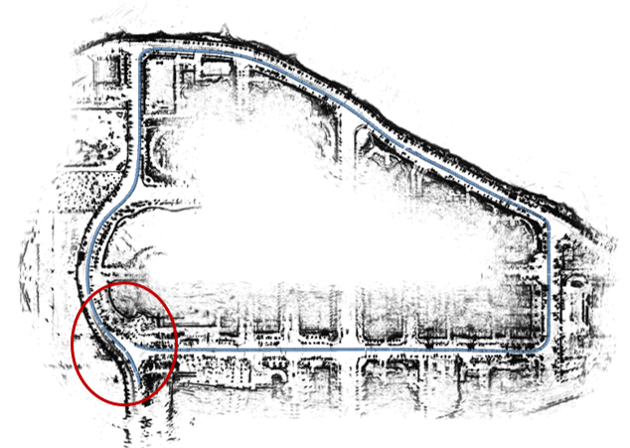}
        \centerline{(b)}
    \end{minipage}
    \begin{minipage}[t]{0.33\linewidth}
        \centering
        \includegraphics[width=\textwidth]{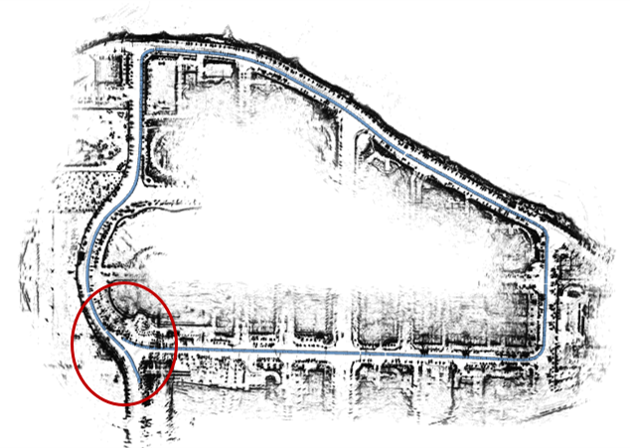}
        \centerline{(c)}
    \end{minipage}
    \caption{\textbf{Visualization of different motion distortion compensation results.} (a) presents the results with motion distortion compensation enabled at all times; (b) shows the results with motion distortion compensation disabled; and (c) displays the outcomes from our strategic motion distortion compensation approach. It is evident that the overlap of point cloud structural features at the location marked by the red circle in (c) is superior to that in (a) and (b).}
    \label{Fig:mc}
\end{figure*}

\section{Experiment} \label{Sec:exper}

\subsection{Experimental Setup}

\subsubsection{\textbf{Dataset}}

We conducted experiments using both public datasets and real-world vehicle tests.

For the public dataset experiments, all sequences are from the MulRan\cite{kimMulRanMultimodalRange2020} and Boreas\cite{burnettBoreasMultiSeasonAutonomous2023} datasets. MulRan is selected for its multi-environment, multi-session, providing radar and IMU data across repeated trajectories, making it well-suited for evaluating robustness in dynamic urban settings. Boreas, which includes diverse weather conditions but is less frequently used, is used for qualitative analysis.

The real-world vehicle tests were carried out in a factory area located in the Shunyi District of Beijing. An experimental vehicle, equipped with a scanning radar, an inertial measurement unit (IMU), and real-time kinematic (RTK) positioning, was employed to gather data for these tests. We have designated these data as “Shunyi”. The aim of the real-world vehicle experiments was to validate the algorithm’s viability for deployment in authentic applications.

\subsubsection{\textbf{Error Metrics}}\label{subSec:error_metrics}

To verify the performance of the odometry, two indicators are selected: odometry accuracy and runtime. The odometry accuracy was evaluated using the classical method introduced in \cite{geigerAreWeReady2012b}, where we recorded and analyzed the relative translation error (\%) and rotation error (°/100m) between trajectory segments of varying lengths (e.g., from 100 meters to 800 meters). As for the running time, it is evaluated based on the duration from receiving the radar data to completing the pose estimation.

\subsubsection{\textbf{Implementation Details}} \label{subSec:Imple}

In Eq.\eqref{Eq:C}, we set the observation to be 6D, rather than the actual 3D observations (2D rotation and 1D yaw angle). This choice was made not only for simplifying the mathematical derivation of the ESKF (e.g., the C matrix can be conveniently written as in Eq.\eqref{Eq:C}), but also to facilitate the implementation of the code. Therefore, when feeding the actual observations into the ESKF, we artificially set $z=0$, $\mathrm{pitch}=\mathrm{roll}=0$, and their covariances were set to 1e-6, indicating that we are only considering pose transformation estimation in the 2D plane.

Noted that the motion distortion compensation in the scanning radar odometry branch of RINO is not always active. Repeated trial results (Fig.\ref{Fig:mc}) indicate that constantly enabling dynamic distortion compensation may actually degrade the system’s rotational estimation accuracy. We imposes thresholds \texttt{min\_for\_mc\_} and \texttt{max\_for\_mc\_} for the motion distortion compensation module. The compensation only activate when the rotation falls between these two threshold values.

Unlike ORORA, we did not perform voxel-sampling on the pointcloud $\lbrace\boldsymbol p_i\rbrace_{i=1}^N$ because we found that this would disrupt the assignment of timestamps to each point.

Furthermore, we observed that when the radar is close to stationary or moving slowly, the computation time increases significantly. This is due to the excessive number of keypoint pairs matched between two similar scenes, which leads to slow MCIS search. To address this, we introduced a threshold \texttt{thr\_for\_stop\_}; if the number of matched point pairs exceeds this threshold, the radar is considered stationary. This may result in some loss of accuracy but ensures stable operation of the radar in real-world scenarios.

\subsubsection{\textbf{Parameters}}

\begin{table}[t]
    \centering
    \caption{system parameter configuration}
    \begin{tabular}{lcc}
        \toprule
        Parameter & Value for MulRan & Value for Shunyi\\
        \midrule
        \texttt{zw\_}&3.0&5.0\\
        \texttt{wf\_}&17&21\\
        \midrule
        \texttt{min\_for\_mc\_}&2.0°&2.0°\\
        \texttt{max\_for\_mc\_}&9.0°&9.0°\\
        \midrule
        \texttt{thr\_for\_stop\_}&600&400\\
        \midrule
        \texttt{thr\_for\_inlier\_}&0.3&0.1\\
        \texttt{cbar\_radial\_}&0.1&0.4\\
        \texttt{cbar\_tangential\_}&0.0262&0.0262\\
        \bottomrule
    \end{tabular}
    \label{Tab:param}
\end{table}

The number of parameters in RINO is small which reflects the model’s strong generalization and adaptability to new data. We listed all the parameters in the Tab.\ref{Tab:param}. The z-value \texttt{zw\_} is used to threshold the noise, and \texttt{wf\_} is the median filter width; both are tunable parameters in the improved cen2018 keypoint extraction module. The meanings of \texttt{min\_for\_mc\_}, \texttt{max\_for\_mc\_}, and \texttt{thr\_for\_stop\_} have been explained earlier in the text. \texttt{thr\_for\_inlier\_} is the threshold used in MCIS to determine whether a point is an inlier. The parameters \texttt{cbar\_radial\_} and \texttt{cbar\_tangential\_} correspond to the values of $\bar{c}^2$ in Eq.\eqref{Eq:rot_est} and Eq.\eqref{Eq:trans_est}. The differences between the two sets of parameters are primarily due to the use of a different scanning radar type. It is important to note that the parameters suitable for the MulRan were obtained through adjustments on the DCC01 sequence and were kept fixed during testing on the remaining sequences, as We believe that this approach to renders the experimental results more meaningful.

\begin{table*}
    \centering
    \caption{Evaluation on 9 sequences of the MulRan dataset using different methods and sensor modalities. All errors are expressed as relative translation error [\%] / relative rotation error [deg /100m].}
    \begin{tabular}{lccccccccc}
        \toprule
        \multirow{2}{*}{Method}&\multicolumn{9}{c}{Sequence}\\
        \cmidrule(lr){2-10} &DCC01&DCC02&DCC03&KAIST01&KAIST02&KAIST03&Riverside01&Riverside02&Riverside03\\
        \midrule
        SuMa~\cite{behleyEfficientSurfelBasedSLAM2018b}&2.71/0.4&4.07/0.9&2.14/0.6&2.90/0.8&2.64/0.6&2.17/0.6&1.66/0.6&1.49/0.5&1.65/0.4\\
        LOAM~\cite{zhangLOAMLidarOdometry2014}&3.16/0.86&2.64/0.74&2.23/0.74&2.70/0.82&2.80/0.84&7.54/0.79&4.25/0.88&4.14/0.90&4.21/1.02\\
        \midrule
        CFEAR-1~\cite{adolfssonLidarLevelLocalizationRadar2022}&2.73/0.73&1.82/0.60&1.77/0.62&2.62/0.97&2.45/0.90&2.85/1.08&2.55/0.90&2.71/0.82&3.56/0.82\\
        CFEAR-2~\cite{adolfssonLidarLevelLocalizationRadar2022}&2.44/0.63&1.65/0.54&1.41/0.50&2.12/0.81&1.93/0.74&2.08/0.87&2.30/0.80&2.07/0.66&2.60/0.59\\
        CFEAR-3~\cite{adolfssonLidarLevelLocalizationRadar2022}&2.28/0.54&1.49/0.46&1.47/0.48&1.59/0.66&1.62/0.66&1.73/0.78&\textcolor{blue}{1.59}/0.83&\textcolor{blue}{1.39}/\textcolor{blue}{0.51}&\textcolor{blue}{1.41}/0.40\\
        CFEAR-3-s50~\cite{adolfssonLidarLevelLocalizationRadar2022}&2.09/0.55&1.38/0.47&1.26/0.47&1.48/0.65&1.51/0.63&\textcolor{blue}{1.59}/0.75&1.62/0.62&\textcolor{red}{1.35}/0.52&\textcolor{red}{1.19}/\textcolor{blue}{0.37}\\
        SDPO~\cite{zhangScanDenoisingNormal2023a}&\textcolor{red}{1.55}/\textcolor{red}{0.35}&1.53/\textcolor{blue}{0.33}&1.60/\textcolor{red}{0.30}&1.57/\textcolor{red}{0.29}&1.61/\textcolor{red}{0.35}&\textcolor{blue}{1.59}/\textcolor{red}{0.32}&1.61/\textcolor{red}{0.26}&1.59/\textcolor{red}{0.27}&1.62/\textcolor{red}{0.29}\\
        $\text{R}^3\text{O}$~\cite{lubancoR3RobustRadon2024}&2.39/\textcolor{blue}{0.43}&1.40/0.34&1.48/0.41&1.89/0.63&1.55/0.53&\textcolor{red}{1.53}/\textcolor{blue}{0.50}&\textcolor{red}{1.34}/\textcolor{blue}{0.39}&1.98/0.53&1.81/0.57\\
        \midrule
        MC-RANSAC~\cite{burnettWeNeedCompensate2021}&4.79/1.27&3.76/1.00&4.39/1.40&-&6.60/1.74& 4.67/1.19&4.76/1.08& 6.49/1.73&7.72/2.19\\
        MC-RANSAC + \texttt{DPLR}~\cite{burnettWeNeedCompensate2021}&4.49/1.12&3.64/1.00&4.37/1.39&-&6.32/1.68& 4.42/1.12&5.92/1.43&8.38/2.08& 6.61/1.88\\
        ORORA(\texttt{Cen2018})~\cite{limORORAOutlierRobustRadar2023a}&3.18/0.66&2.38/0.57&2.77/0.79&-&3.12/0.78&2.53/0.57&3.51/0.76&3.31/0.76&2.79/0.64\\
        ORORA(\texttt{Cen2019})~\cite{limORORAOutlierRobustRadar2023a}&3.12/0.67&2.60/0.51&2.37/0.57&-&3.28/0.82&3.04/0.70&3.53/0.84&2.67/0.64&2.11/0.49\\
        RINO(\texttt{Cen2018})&\textcolor{blue}{1.64}/0.46&\textcolor{blue}{1.25}/\textcolor{red}{0.30}&\textcolor{red}{0.91}/0.39&\textcolor{blue}{1.15}/\textcolor{blue}{0.43}&\textcolor{blue}{1.42}/0.46&2.08/1.08&2.28/0.53&2.35/0.80&2.28/0.47\\
        RINO(\texttt{Cen2019})&1.71/0.49&\textcolor{red}{1.23}/\textcolor{blue}{0.33}&\textcolor{blue}{1.08}/\textcolor{blue}{0.35}&\textcolor{red}{1.08}/\textcolor{blue}{0.43}&\textcolor{red}{1.32}/\textcolor{blue}{0.45}&1.87/0.96&2.40/0.50&2.11/0.64&2.20/0.48\\
        \bottomrule
    \end{tabular}
    \label{Tab:benchmarks}
\end{table*}

\begin{figure*}
    \centering
    \includegraphics[width=\linewidth]{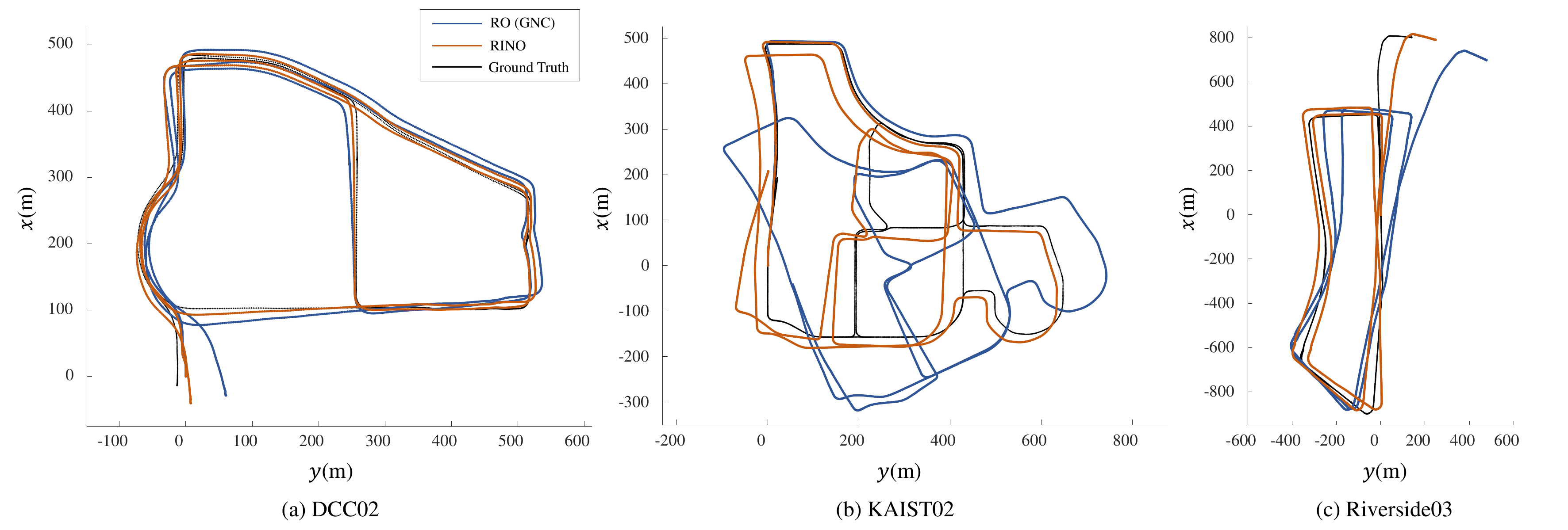}
    \caption{\textbf{Intuitive comparison between RINO and the baseline.} The black trajectory represents the ground truth, the blue trajectory represents the trajectory estimated by the baseline, and the orange trajectory represents the trajectory estimated by RINO.}
    \label{Fig:mulran_odom}
\end{figure*}

\begin{figure}
    \centering
    \includegraphics[width=\linewidth]{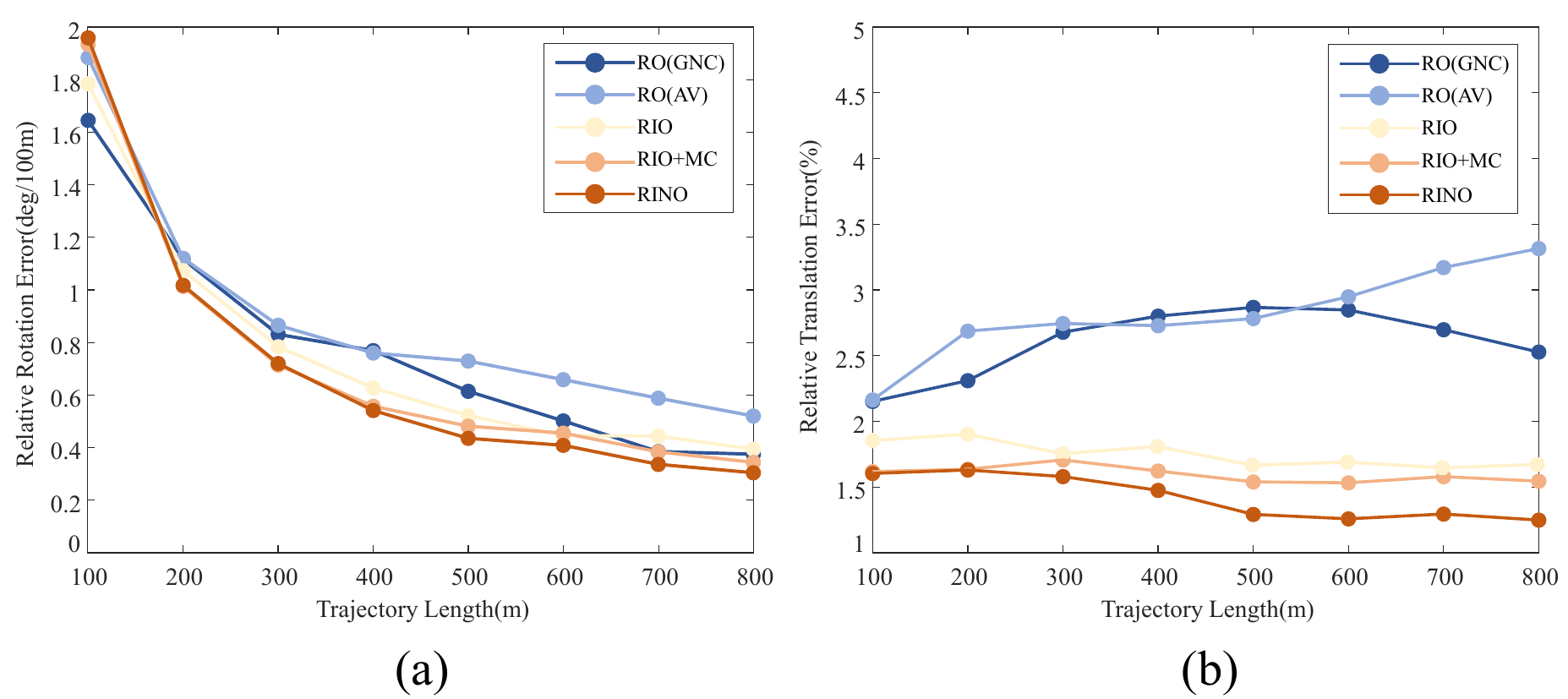}
    \caption{\textbf{Comparison of drift with respect to path distance on the DCC02 sequence of MulRan dataset.} (a) Rotation error. (b) Translation error.}
    \label{Fig:mulran_error}
\end{figure}

\subsection{Benchmark Results Within MulRan Dataset}

We selected the best odometry or SLAM algorithms using different sensor modalities in large-scale environments as benchmarks. These methods include LOAM~\cite{zhangLOAMLidarOdometry2014}, SuMa~\cite{behleyEfficientSurfelBasedSLAM2018b}, and the state-of-the-art scanning radar odometry algorithms introduced in Sec.\ref{Sec:Relat}: CFEAR~\cite{adolfssonLidarLevelLocalizationRadar2022}, SDPO~\cite{zhangScanDenoisingNormal2023a}, $\text{R}^3\text{O}$~\cite{lubancoR3RobustRadon2024}, MC-RANSAC~\cite{burnettWeNeedCompensate2021}, and ORORA~\cite{limORORAOutlierRobustRadar2023a}. Specifically, CFEAR is evaluated in four configurations (CFEAR-1/2/3/3-s50) differing in solving strategy, keyframe settings, and robustness tuning. MC-RANSAC estimates motion with distortion compensation, while MC-RANSAC+\texttt{DPLR} further handles Doppler distortion. ORORA is tested with different feature extraction methods (Cen2018 and Cen2019) to assess the impact of feature quality.

We conducted tests on nine sequences from the MulRan dataset: DCC01, DCC02, DCC03, KAIST01, KAIST02, KAIST03, Riverside01, Riverside02, and Riverside03. The Sejong sequences were excluded from evaluation, as they are designed for global localization rather than odometry, due to their large structural variations. The detailed results are shown in Tab.\ref{Tab:benchmarks}. In the table, the red data points represent the minimum errors, while the blue data points correspond to the second smallest errors. It is recommended to view the table in color for improved clarity. It is important to note that, due to the unavailability of direct reproductions of these methods, the data presented in the table is taken from the reported results in publicly published papers. The table categorizes the aforementioned methods into LiDAR odometry, state-of-the-art radar odometry methods, and the baselines. The data is presented in the format (\%/°/100m), where the first value represents the relative translation error, and the second value represents the relative rotation error.

The proposed method, RINO, demonstrates relatively small relative translation and rotation errors across the majority of sequences, with particularly impressive performance in the DCC and KAIST sequences. RINO significantly improves upon the baseline, ORORA. Specifically, compared to ORORA with Cen2018 and Cen2019 keypoint extraction, RINO achieves average reductions of 1.43\% and 1.73\% in translation error, and 0.20°/100m and 0.13°/100m in rotation error in the DCC and KAIST sequences, respectively. In the DCC03 sequence, the error reduction reaches up to 1.46\% in translation and 0.28°/100m in rotation. Moreover, when compared to LiDAR-based methods such as LOAM and SuMa, which offer higher resolution and suffer less from per-frame motion distortion, RINO has demonstrated superior performance. We conclude that RINO has achieved a level of radar odometry performance comparable to the state-of-the-art.

When RINO is tested on the Riverside sequences, there is a notable decline in odometry accuracy compared to the DCC and KAIST sequences. The Riverside sequences, which were collected along a riverbank, represent a degenerate scenario that challenges odometry systems. Furthermore, the performance degradation of RINO is more pronounced in these sequences compared to other algorithms. A possible explanation for this lies in the Markovian assumption inherent in our pipeline. Both the Extended Kalman Filter (EKF) and the odometry based on global point cloud registration operate in an incremental fashion, where the current pose transformation estimate is solely dependent on the previous time step. This approach, to some extent, overlooks the broader structural information of point clouds in the local map, potentially leading to confusion in degenerate environments.

\subsection{Ablation Studies Within MulRan Dataset}

To better highlight the effects of each improvement we proposed, we conducted multiple ablation experiments on 9 sequences from the MulRan dataset. The results are shown in Tables 1, 2, and 3. The data under the DCC, KAIST, and Riverside categories represent the average values of sequences 01, 02, and 03. All results are presented in the format (\%/°/100m).

\begin{table}
    \centering
    \caption{Odometry results for different pointcloud registration methods. All errors are expressed as relative translation error [\%] / relative rotation error [deg /100m].}
    \begin{tabular}{lccc}
        \toprule
         \multirow{2}{*}{Method}&\multicolumn{3}{c}{Sequence}\\
         \cmidrule(lr){2-4} &DCC&KAIST&Riverside\\
        \midrule
        RO (Graduated Non-Convexity)&2.94/0.62&4.14/0.94&5.40/0.81\\
        RO (Adaptive Voting)&2.93/0.59&4.71/1.04&5.16/0.85\\
        \bottomrule
    \end{tabular}
    \label{Tab:gnc_av}
\end{table}

\subsubsection{\textbf{Effectiveness of Adaptive Voting}}

We implemented a scanning radar odometry method, RO(GNC), using progressive non-convex optimization to solve pointcloud registration problem, which represents ORORA receiving sensor data via ROS (Robot Operating System). For comparison, we also implemented RO(AV) using adaptive voting to solve pointcloud registration, which represents the scanning radar branch in RINO. Both RO(GNC) and RO(AV) employ the improved keypoint extraction technique described in Sec.~\ref{subSec:Radar}, which differs from the code released by ORORA. Before feature matching, no voxel grid filtering is applied to the pointcloud in either method. This decision was based on our observations during the replication process, where we found that performing this step disrupted the timestamps assigned sequentially to each keypoint, thereby invalidating the motion distortion compensation. RO(GNC) performance is degraded compared to the test results presented in the ORORA paper (Tab.\ref{Tab:benchmarks}). This degradation is hypothesized to arise from differences in parameter settings and the loss of precise azimuth encoding information in the raw scanning radar data. The comparison results between the two methods are shown in Tab.\ref{Tab:gnc_av}. The odometry errors of RO(GNC) and RO(AV) are similar across multiple sequences. However, in subsequent runtime tests (Fig.\ref{Fig:runtime}), we found that the running time of RO(AV) is generally shorter than that of RO(GNC). Therefore, we conclude that the adaptive voting method improves the system's efficiency.

\begin{table}
    \centering
    \caption{Odometry results for different coupling settings. All errors are expressed as relative translation error [\%] / relative rotation error [deg /100m].}
    \begin{tabular}{lccccc}
        \toprule
         \multirow{2}{*}{Method}&\multicolumn{2}{c}{Coupling}&\multicolumn{3}{c}{Sequence} \\
         \cmidrule(lr){2-3} \cmidrule(lr){4-6} &IMU&Uncertainty&DCC&KAIST&Riverside\\
        \midrule
        RO&-&-&2.93/0.59&4.71/1.04&5.16/0.85\\
        RIO&\checkmark&-&1.80/0.47&1.98/0.74&3.79/0.70\\
        \cellcolor{gray!20}RIO*&\checkmark&\checkmark&\cellcolor{gray!20}\textbf{1.67}/\textbf{0.46}&\cellcolor{gray!20}\textbf{1.81}/\textbf{0.67}&\cellcolor{gray!20}\textbf{3.54}/\textbf{0.68}\\
        \bottomrule
    \end{tabular}
    \label{Tab:rio}
\end{table}

\subsubsection{\textbf{Effectiveness of Adaptive Coupling}}

We extends the RO(AV) model by integrating an IMU to establish a loosely-coupled odometry system. In this configuration, the scanning radar subsystem does not explicitly compute the uncertainty of the pose transformation estimate during pointcloud registration. Therefore, during the ESKF observation process, the covariance of the observations from the scanning radar branch is fixed to $10^{-2}$. This model is denoted as RIO in Tab.\ref{Tab:rio}. Subsequently, we implemented an adaptive coupling system that adjusts the fusion weights based on the uncertainty estimated from the scanning radar branch, referred to as RIO* in Tab.\ref{Tab:rio}. Compared to RO, RIO exhibits a significant performance improvement across all sequences, highlighting that loosely coupling the IMU with the scanning radar odometry substantially enhances the system's accuracy. Additionally, incorporating the uncertainty in the pose estimation from the scanning radar branch further reduces odometry errors. This adjustment allows RINO to adapt based on the sensor state, enabling real-time adaptive adjustments. Such adaptive capability is crucial for handling extreme situations more effectively, including sensor failures or scene degradation.

\begin{table}
    \centering
    \caption{Odometry results for different motion compensation strategies. All errors are expressed as relative translation error [\%] / relative rotation error [deg /100m].}
    \begin{tabular}{lcccc}
        \toprule
         \multirow{2}{*}{Method}&\multirow{2}{*}{Motion Compensation}&\multicolumn{3}{c}{Sequence} \\
         \cmidrule(lr){3-5} &&DCC&KAIST&Riverside\\
        \midrule
        $\text{RIO}^{\ddag}$&on&2.56/0.52&3.59/0.91&4.12/0.98\\
        $\text{RIO}^{\dag}$&off&1.67/0.46&1.81/0.67&3.54/0.68\\
        \cellcolor{gray!20}RINO&strategic&\cellcolor{gray!20}\textbf{1.27}/\textbf{0.38}&\cellcolor{gray!20}\textbf{1.55}/\textbf{0.66}&\cellcolor{gray!20}\textbf{2.30}/\textbf{0.60}\\
        \bottomrule
    \end{tabular}
    \label{Tab:mc}
\end{table}

\begin{table}
    \centering
    \caption{Component-wise runtime for RIO(GNC) and RIO(AV). All times are reported in seconds [s].}
    \label{Tab:runtime}
    \begin{tabular}{lcc}
    \toprule
    Module & RIO(GNC) & RIO(AV) \\
    \midrule
    \underline{Keypoints Extraction \& Matching} & 0.0942\scriptsize{$\pm$ 0.0169} & 0.0920\scriptsize{$\pm$ 0.0166} \\
    IMU Prediction                   & 0.0001\scriptsize{$\pm$ 0.0000} & 0.0001\scriptsize{$\pm$ 0.0000} \\
    Motion Undistortion             & 0.0004\scriptsize{$\pm$ 0.0002} & 0.0004{\scriptsize$\pm$ 0.0002} \\
    \cellcolor{gray!20}\underline{Pose Estimation \& Fusion}       & 0.0507\scriptsize{$\pm$ 0.0149} & \textcolor{red}{0.0452}\scriptsize{$\pm$ \textcolor{red}{0.0110}} \\
    Pointcloud Map Publishing       & 0.0004\scriptsize{$\pm$ 0.0011} & 0.0004\scriptsize{$\pm$ 0.0010} \\
    \midrule
    \textbf{Total}                  & \textbf{0.1458} & \textbf{0.1381} \\
    \bottomrule
    \end{tabular}
\end{table}

\subsubsection{\textbf{Effectiveness of motion compensation strategy}}

Earlier, the qualitative effects of different motion distortion compensation strategies were demonstrated in Fig.\ref{Fig:mc}. In this section, we perform a quantitative evaluation (Tab.\ref{Tab:mc}) of these strategies, incorporating the engineering adjustments described in Sec.\ref{subSec:Imple}. $\text{RIO}^{\ddag}$ refers to the model with continuous motion compensation, $\text{RIO}^{\dag}$ represents the model with motion compensation disabled, and RINO denotes our final model with the adaptive motion compensation strategy. Experimental results indicate that keeping motion compensation continuously enabled actually detracts from the model's performance. We attribute this to the relatively low scanning frequency of the scanning radar. Within a single scanning cycle (0.25 seconds), significant drift in the gyro bias of the IMU occurs, and using this drift for motion distortion compensation may negatively impact rotation estimation. We hypothesize that motion compensation may perform better during large-angle rotations, where the influence of gyro bias on angular velocity is comparatively smaller. The performance improvement of RINO over both $\text{RIO}^{\ddag}$ and $\text{RIO}^{\dag}$ partially supports this hypothesis.

\subsubsection{\textbf{Visualization}}

We present the estimated trajectories in Fig.\ref{Fig:mulran_odom}, and the path error in Fig.\ref{Fig:mulran_error}. In Fig.\ref{Fig:mulran_odom}, the trajectory estimated by RINO clearly aligns much more closely with the ground truth trajectory. In Fig.\ref{Fig:mulran_error}, RIO represents the loosely coupled radar-IMU system, while RINO represents the adaptive coupling system between the two.

\begin{figure}
    \includegraphics[width=\linewidth]{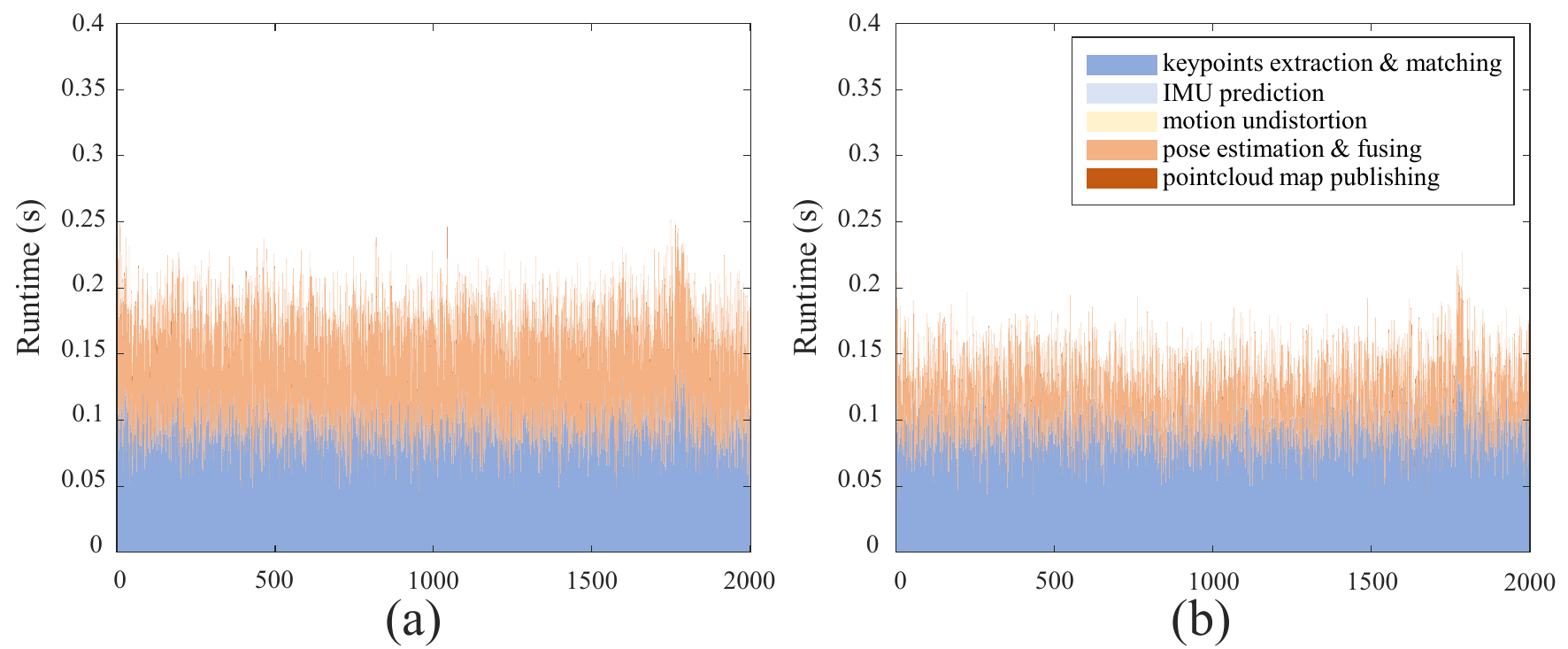}
    \caption{\textbf{Runtime comparison of each modules} in (a) RO(GNC)  and (b) RO(AV) on the DCC01 sequence of the MulRan dataset.}
    \label{Fig:runtime}
\end{figure}

\subsection{Runtime}

RINO is capable of real-time operation, as demonstrated in Fig.\ref{Fig:runtime}. The algorithm's runtime test was conducted using the DCC01 sequence from the MulRan dataset. Our hardware setup consists of an Intel® Core i5-13600K processor.

We provide a detailed breakdown of the runtime for each module in RINO, including keypoint extraction and matching, IMU prediction, motion undistortion, pose estimation and fusion, and point cloud map publishing. Similar to ORORA, the majority of the system's runtime is spent on keypoint extraction and matching. Notably, the runtime of the point cloud registration module is influenced by the number of keypoint correspondences, and its efficiency tends to degrade when the vehicle is moving slowly or remains stationary.

While a direct runtime comparison with ORORA is desirable, it is not presented due to differences in execution environments and data pipelines. Specifically, ORORA operates offline by reading pre-recorded data, whereas RINO simulates real-time operation via ROS message publishing, introducing additional system overhead. To provide a more meaningful performance comparison, we instead analyze the runtime of RIO(GNC)—which adopts the same robust optimization strategy as ORORA—and RIO(AV), which incorporates our proposed Adaptive Voting approach. This comparison offers insight into the computational efficiency improvements achieved by our enhancements.

As shown in Tab.\ref{Tab:runtime}, the average runtime of RINO is approximately 0.1381s per frame, ensuring real-time performance at the radar acquisition frequency of 4Hz. Furthermore, Fig.\ref{Fig:runtime} compares RIO(GNC) and RIO(AV), and shows that the adoption of the Adaptive Voting strategy results in an 11.5\% improvement in the efficiency of pointcloud registration. In addition, the reduced variance observed in RIO(AV) indicates that the Adaptive Voting solver exhibits greater stability compared to the GNC-based approach.

\subsection{Discussion about Estimation Uncertainty}

\begin{figure}
    \centering
    \includegraphics[width=\linewidth]{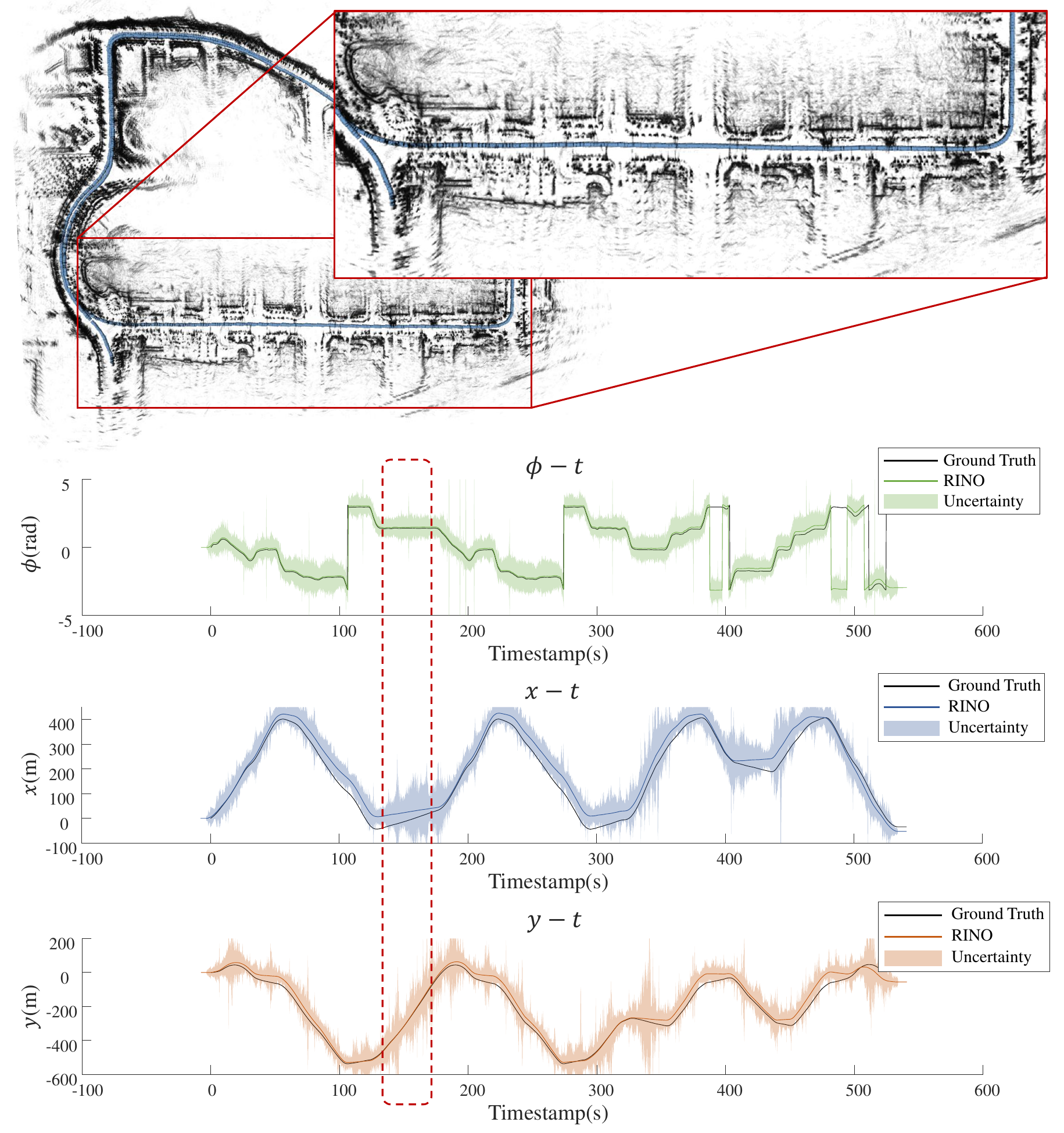}
    \caption{\textbf{Relation between uncertainty and scene.} The pose and uncertainty are obtained from the scanning radar branch on the DCC01 sequence of the MulRan dataset. The uncertainty is represented by scaling the variance of the pose estimate. The red boxes highlight the correspondence between the data in the figure and the vehicle's position in the scene.}
    \label{Fig:scene_uncertainty}
\end{figure}

\begin{figure}
    \centering
    \includegraphics[width=\linewidth]{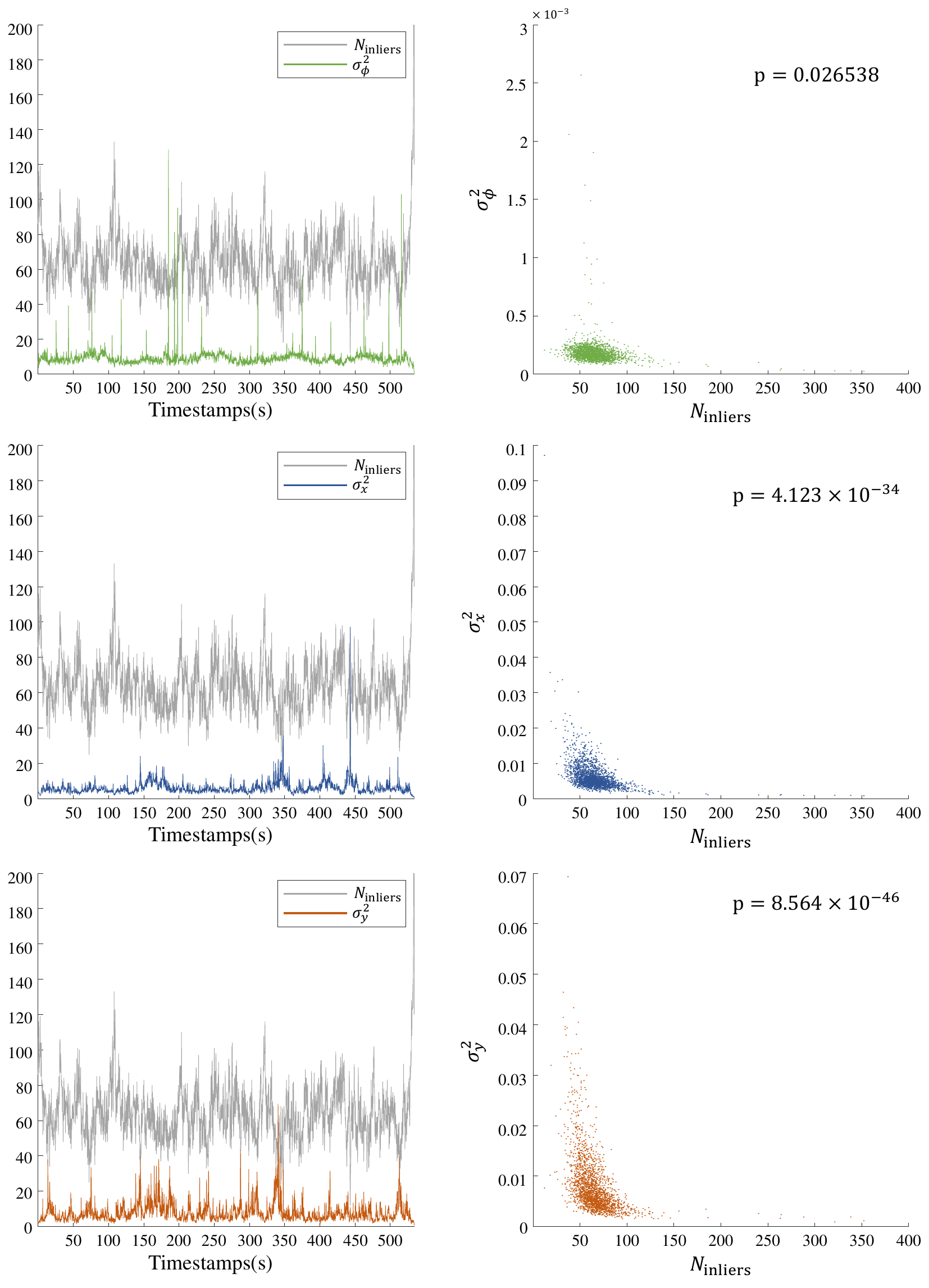}
    \caption{\textbf{Relation between uncertainty and number of inliers.} The uncertainty and the number of inliers obtained from the scanning radar branch on the DCC01 sequence of the MulRan dataset. $N_{\text{inliers}}$ means the number of points in the inlier set remaining after Maximum Clique Inlier Selection (MCIS) and the uncertainty of the pose estimate. As the number of inlier points increases, the uncertainty in the pose estimate generally decreases, reflecting higher confidence in the alignment. Conversely, a smaller inlier set leads to higher uncertainty, indicating less reliable pose estimation.}
    \label{Fig:inliers_uncertainty}
\end{figure}

First, we discuss the relationship between the uncertainty in radar pose estimation and the distribution of scene features. In Fig.\ref{Fig:scene_uncertainty}, the red boxes indicate areas with higher uncertainty, corresponding to the vehicle traveling along a straight road from right to left. The building features on both sides of the road are distributed along the road and form a degenerate structure along the y-axis, resulting in higher uncertainty in the y-direction. This observation suggests that pose estimation uncertainty is closely related to the distribution of features in the scene.

Next, we consider the effect of the number of inliers remaining after the maximum clique inlier selection (MCIS) in point cloud registration on pose estimation uncertainty. Intuitively, fewer inliers lead to lower confidence in the estimation. Fig.\ref{Fig:inliers_uncertainty} confirms this, where the troughs in the number of inliers generally coincide with the peaks in pose uncertainty. Furthermore, the smaller p-values provide additional evidence of a direct relationship between the number of inliers and the confidence in pose estimation.

\subsection{Qualitative Evaluation Within Boreas}

\begin{figure}
    \begin{minipage}[t]{\linewidth}
        \centering
        \includegraphics[width=\textwidth]{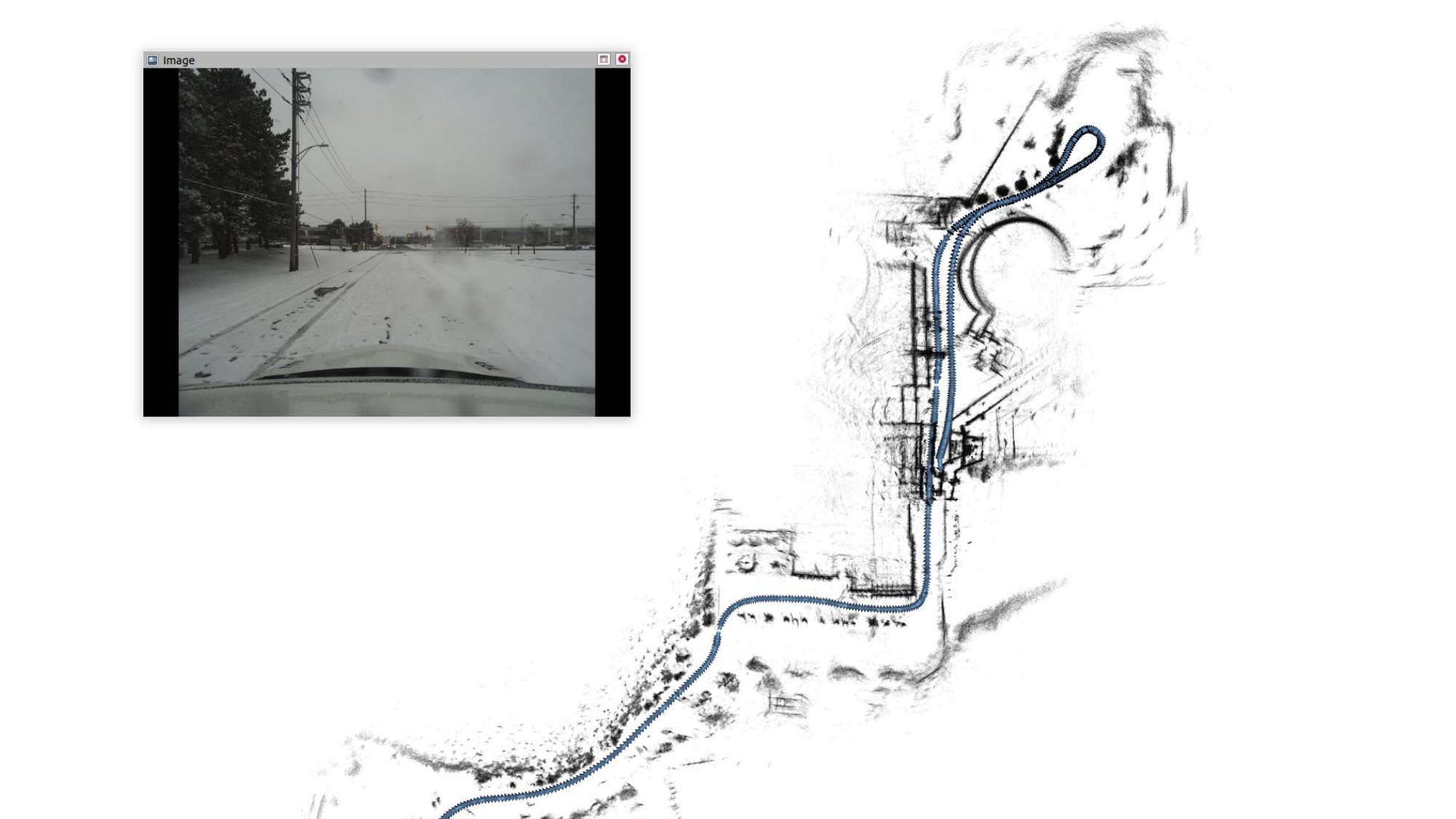}
        \centerline{(a)}
    \end{minipage}
    \begin{minipage}[t]{\linewidth}
        \centering
        \includegraphics[width=\textwidth]{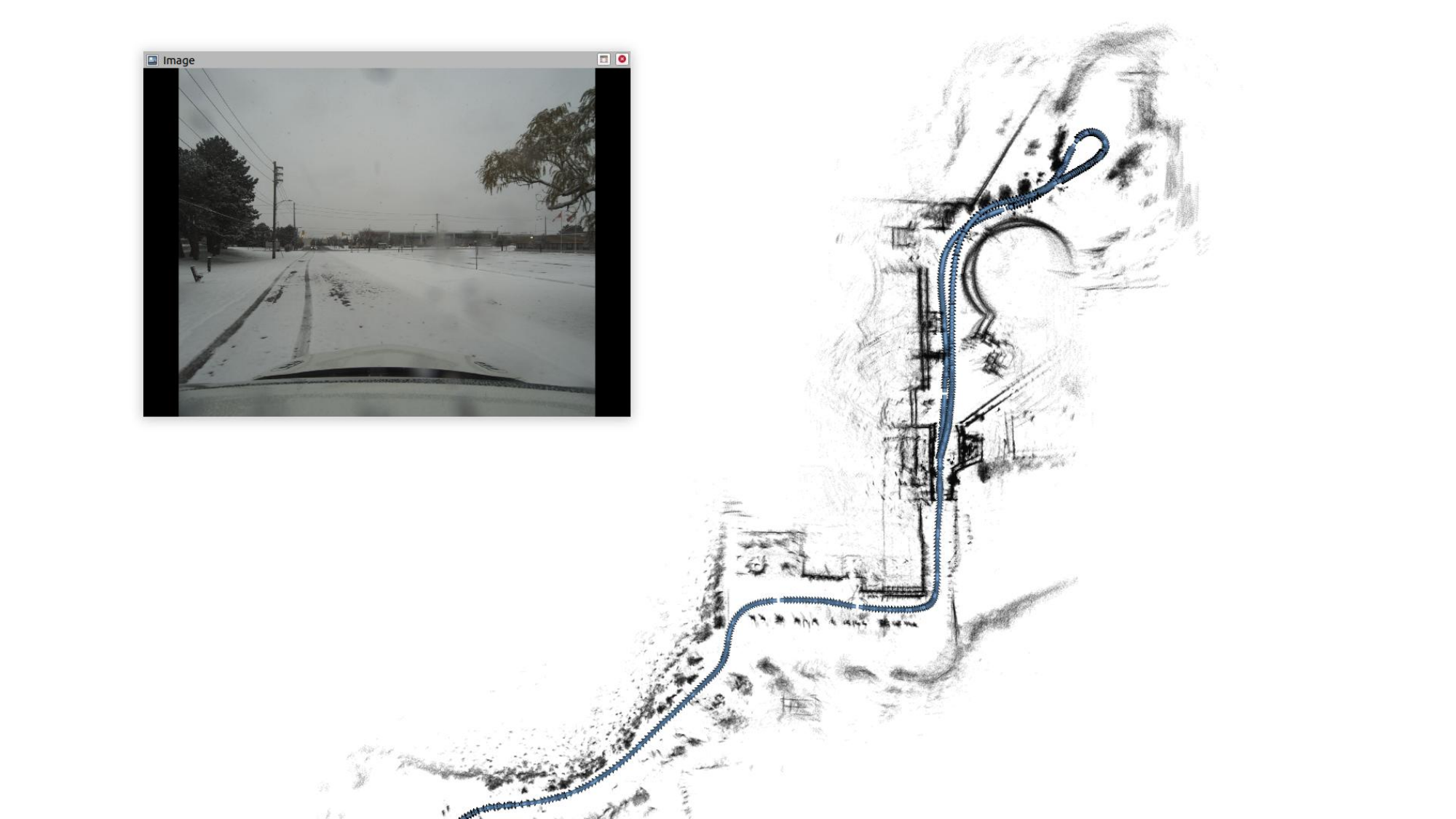}
        \centerline{(b)}
    \end{minipage}
    \begin{minipage}[t]{\linewidth}
        \centering
        \includegraphics[width=\textwidth]{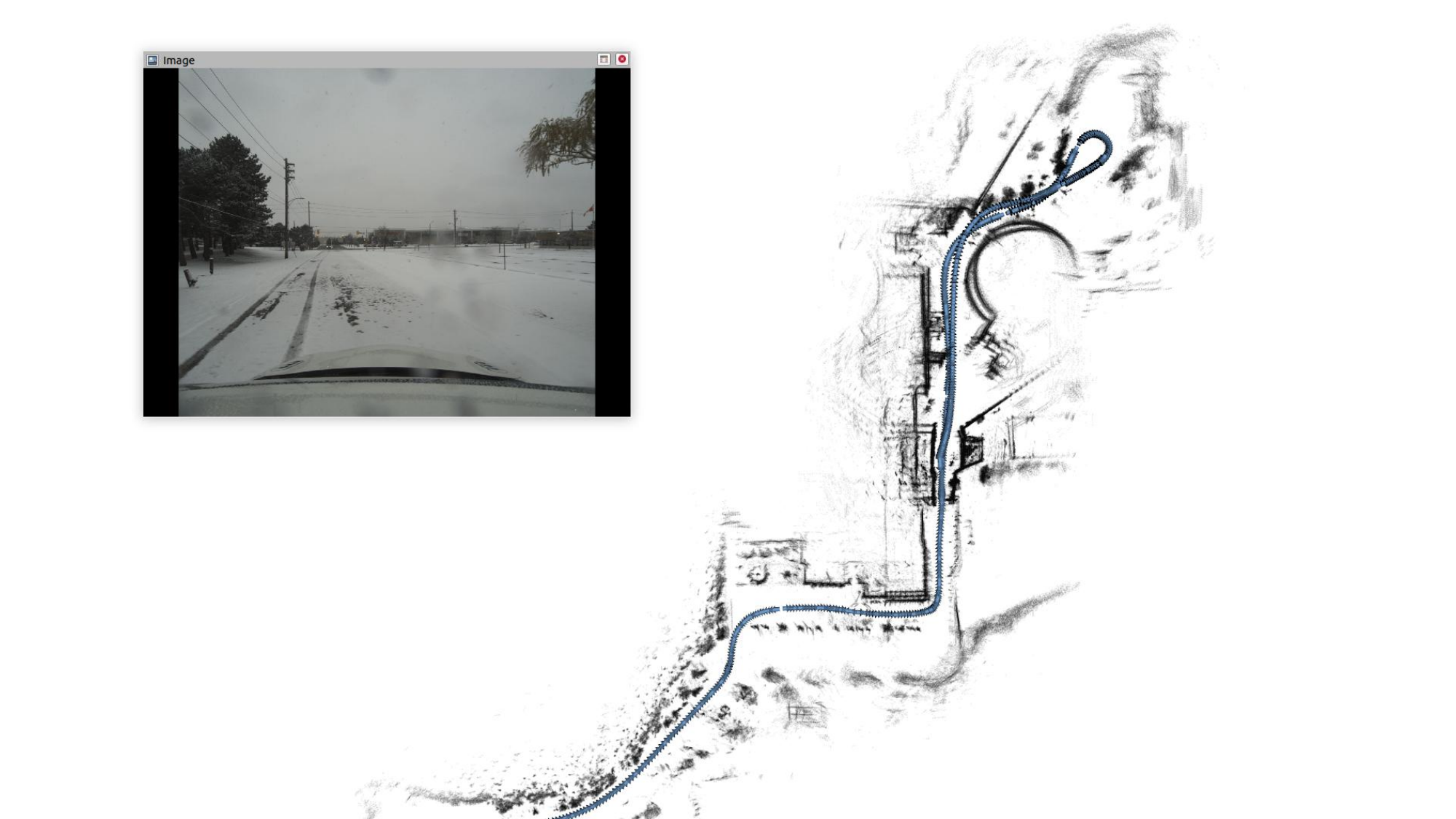}
        \centerline{(c)}
    \end{minipage}
    \caption{\textbf{Visualization of estimated trajectory (blue arrows) and pointcloud map (black points) on the Boreas dataset.} (a) RO: IMU is disabled. After the vehicle turns around in a narrow area, the point cloud in the revisited region is not aligned. (b) RIO: Radar and IMU are loosely coupled. The introduction of IMU improves odometry performance, and the point cloud in the revisited area becomes relatively clearer. (c) RINO: Further implementation of adaptive coupling between radar and IMU. The point cloud in the revisited area is almost perfectly aligned with the point cloud in (b).}
    \label{Fig:boreas}
\end{figure}

We present the performance of RINO on the 2021-11-28-09-18 sequence of Boreas dataset, as shown in Fig.\ref{Fig:boreas}. 

Fig.\ref{Fig:boreas} illustrates the synchronized image data captured by the camera. From the camera's perspective, it is evident that the current weather conditions are harsh, characterized by dim lighting, snow-covered ground, and falling snowflakes that partially obscure the camera's view. The weak illumination and uniformly white snow tend to blur texture features in the camera images, while the floating snowflakes introduce significant noise to the LiDAR scans. These challenging conditions negatively impact both visual and LiDAR-based odometry algorithms.

The experimental results show that RINO remains largely unaffected by these adverse weather conditions.

\subsection{Real-world Evaluation}

We validated the feasibility and robustness of RINO by deploying it on real vehicles operating in previously unseen environments.

\begin{figure}
    \centering
    \includegraphics[width=\linewidth]{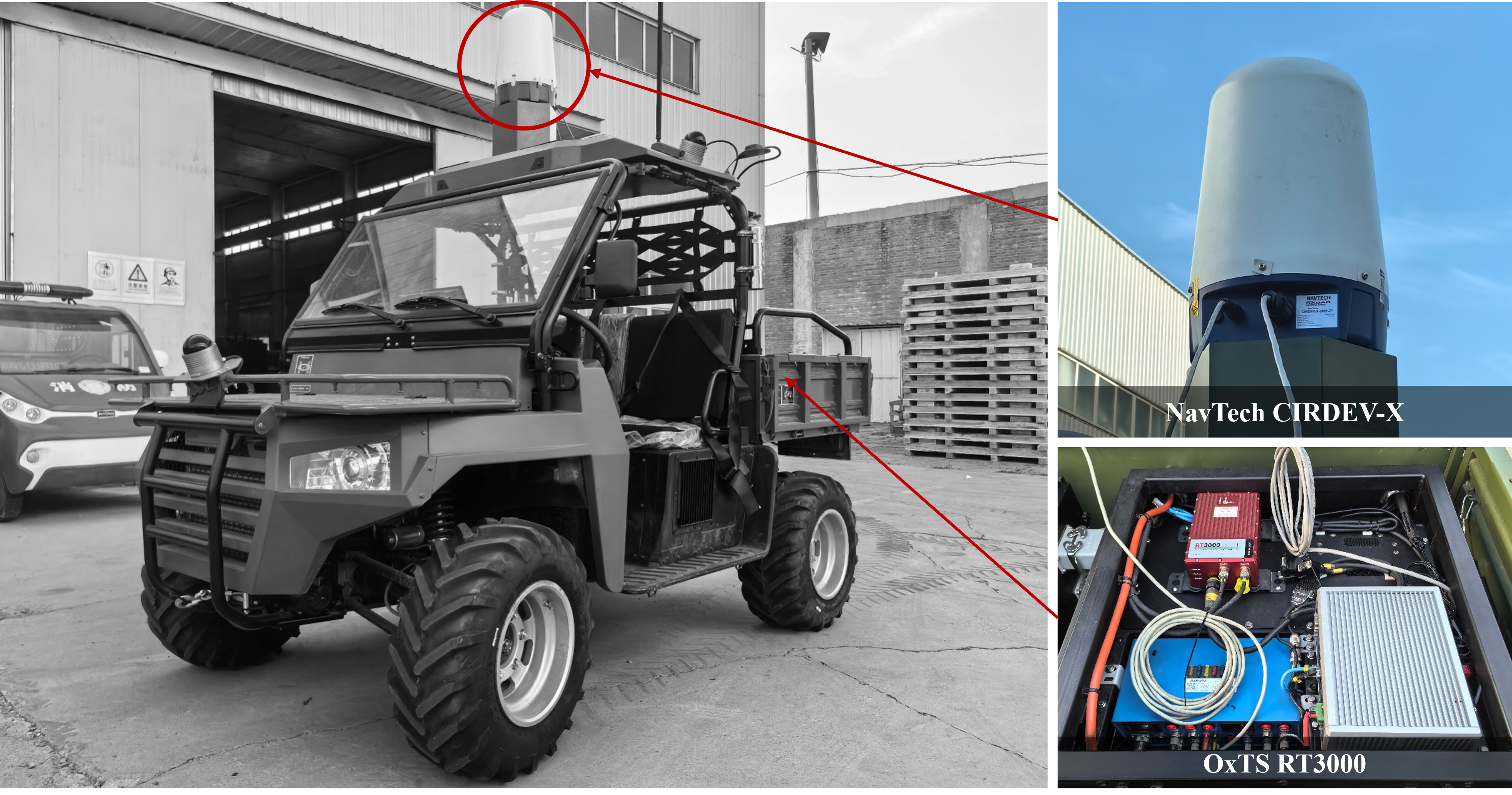}
    \caption{\textbf{Sensor system for real-world experiments.} The sensor system is equipped with a scanning radar and an IMU, which provides the conditions for the fusion of data from both devices.}
    \label{Fig:wild_cat}
\end{figure}

The data collection vehicle is a 4×4 off-road platform equipped with a range of sensors to support various perception tasks, as depicted in Fig.\ref{Fig:wild_cat}. An OxTS RT3000 high-precision positioning system, which integrates an IMU, is installed on the vehicle. A NavTech CIRDEV-X scanning radar is centrally mounted on the roof. For brevity, details of the other sensors are omitted.

The experiment was conducted within a factory in Shunyi District, Beijing. The data collection vehicle followed two distinct routes. The first, a shorter route, was used for parameter tuning (labeled 2024-05-28-17-09), while the second, a longer closed-loop route, was employed for qualitative evaluation of the fusion localization and mapping system (labeled 2024-05-28-18-48). During the experiment, data from both the scanning radar and IMU were transmitted to an industrial computer in ROS message format.

\begin{figure}
    \centering
    \includegraphics[width=\linewidth]{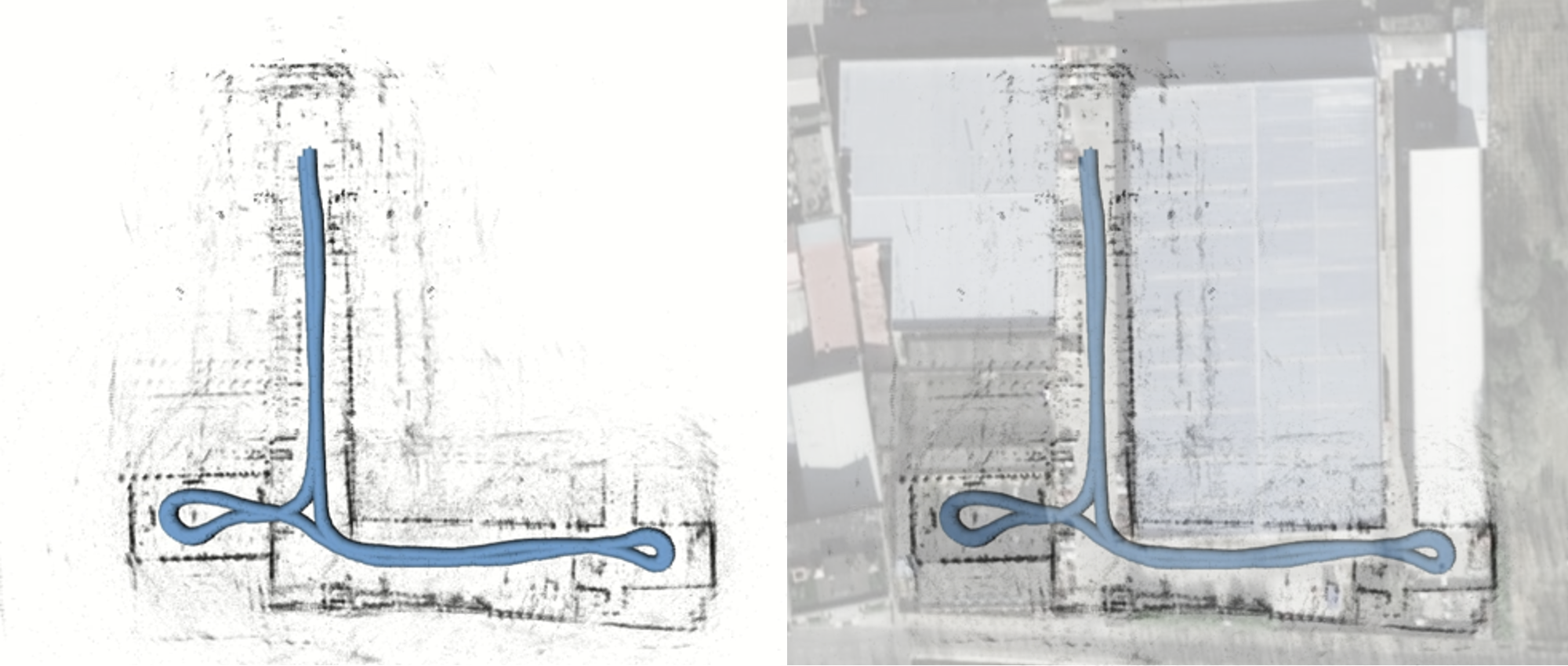}
    \caption{\textbf{Visualization of estimated trajectory (blue arrows) and pointcloud map (black points) in real-world experiments.} The trajectory and map are well aligned with the satellite imagery, demonstrating good accuracy.}
    \label{Fig:results_ours}
\end{figure}

\begin{figure}
    \centering
    \includegraphics[width=\linewidth]{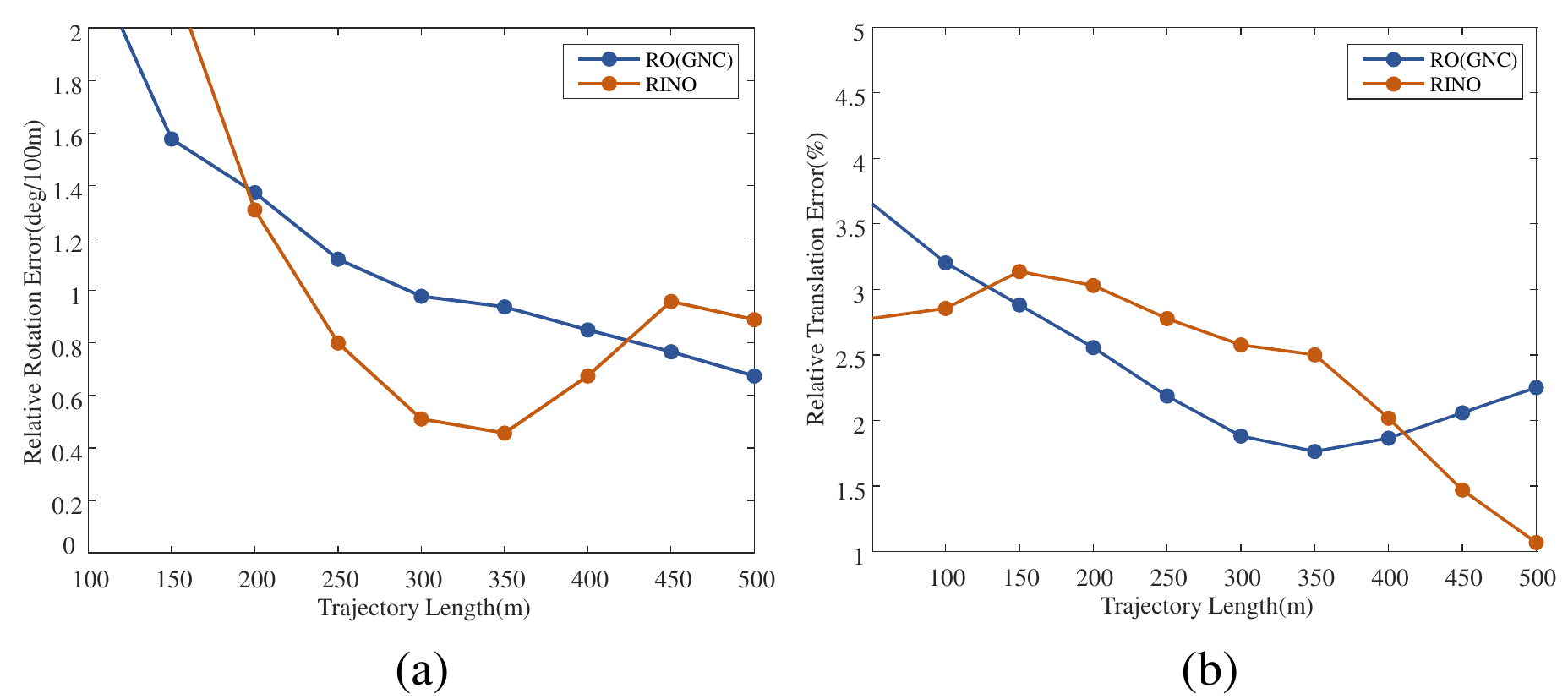}
    \caption{\textbf{Comparison of drift with respect to path distance in real-world experiments.} (a) Rotation error. (b) Translation error.}
    \label{Fig:errors_ours}
\end{figure}

It is important to note that the scanning radar used in our setup differs from those mounted on vehicles in the MulRan or Boreas datasets, as it features a lower radial resolution (only 0.175m). The original images from the NavTech CIRDEV-X radar exhibit more blurred environmental features and contain considerably more noise. Despite these challenges, the experimental results demonstrate that our algorithm remains robust even when processing such degraded data, highlighting the critical role of parameter tuning.

Furthermore, the platform is powered by a diesel engine and equipped with a stiff suspension system, which induces significant vehicle body vibrations during operation. To mitigate IMU measurement errors caused by these vibrations, we implemented a Chebyshev filter. This filtering module has substantially improved the reliability and accuracy of the IMU data.

As shown in Fig.\ref{Fig:results_ours}, the pointcloud map is relatively clear in the revisited regions of both loops. We also present the path error in Fig.\ref{Fig:errors_ours}. These indicates that, even under complex and aggressive motion conditions, our algorithm maintains high-precision odometry estimation, with the pointcloud map faithfully reflecting the structural features present in the satellite map.
\section{Conclusion} \label{Sec:concl}

In this study, we introduce \textbf{RINO}, a robust and accurate Radar-Inertial Odometry framework designed to address the challenges of autonomous navigation for self-driving vehicles in severe weather conditions. By employing a non-iterative approach for rotation and translation estimation, RINO enhances computational efficiency while maintaining stable performance. Key innovations of RINO include an adaptive voting scheme for 2D rotation estimation and the integration of scanning radar pose uncertainty into the Error-State Kalman Filter (ESKF), both of which contribute to improved robustness and accuracy in pose estimation.

Our experimental results demonstrate that RINO outperforms existing state-of-the-art methods and baseline systems in terms of both accuracy and resilience. Notably, strategic motion distortion compensation, the elimination of iterative solving processes, and the incorporation of pose uncertainty estimation play a crucial role in the algorithm's success. Furthermore, extensive real-world testing confirms the system's reliability and robustness, validating its suitability for practical deployment in autonomous driving scenarios.

This work highlights the potential of combining radar and IMU data within a non-iterative, loosely coupled framework and underscores the significance of accurately modeling uncertainty in pose estimation. While RINO achieves superior performance relative to existing approaches, it also identifies areas for further development, particularly in refining motion distortion compensation and addressing degraded operational scenarios. Future research will focus on overcoming these limitations to further enhance the algorithm's robustness and applicability in real-world autonomous navigation contexts.




\bibliographystyle{IEEEtran}
\bibliography{main}

\begin{thebibliography}{10}
\providecommand{\url}[1]{#1}
\csname url@samestyle\endcsname
\providecommand{\newblock}{\relax}
\providecommand{\bibinfo}[2]{#2}
\providecommand{\BIBentrySTDinterwordspacing}{\spaceskip=0pt\relax}
\providecommand{\BIBentryALTinterwordstretchfactor}{4}
\providecommand{\BIBentryALTinterwordspacing}{\spaceskip=\fontdimen2\font plus
\BIBentryALTinterwordstretchfactor\fontdimen3\font minus \fontdimen4\font\relax}
\providecommand{\BIBforeignlanguage}[2]{{%
\expandafter\ifx\csname l@#1\endcsname\relax
\typeout{** WARNING: IEEEtran.bst: No hyphenation pattern has been}%
\typeout{** loaded for the language `#1'. Using the pattern for}%
\typeout{** the default language instead.}%
\else
\language=\csname l@#1\endcsname
\fi
#2}}
\providecommand{\BIBdecl}{\relax}
\BIBdecl

\bibitem{qinVINSMonoRobustVersatile2018b}
T.~Qin, P.~Li, and S.~Shen, ``{{VINS-Mono}}: {{A Robust}} and {{Versatile Monocular Visual-Inertial State Estimator}},'' \emph{IEEE Transactions on Robotics}, vol.~34, no.~4, pp. 1004--1020, Aug. 2018.

\bibitem{camposORBSLAM3AccurateOpenSource2021b}
C.~Campos, R.~Elvira, J.~J.~G. Rodriguez, J.~M. M.~Montiel, and J.~D.~Tardos, ``{{ORB-SLAM3}}: {{An Accurate Open-Source Library}} for {{Visual}}, {{Visual}}--{{Inertial}}, and {{Multimap SLAM}},'' \emph{IEEE Transactions on Robotics}, vol.~37, no.~6, pp. 1874--1890, Dec. 2021.

\bibitem{10287884}
W.~Guan, P.~Chen, Y.~Xie, and P.~Lu, ``Pl-evio: Robust monocular event-based visual inertial odometry with point and line features,'' \emph{IEEE Transactions on Automation Science and Engineering}, vol.~21, no.~4, pp. 6277--6293, 2024.

\bibitem{10488029}
H.~Jiang, R.~Qian, L.~Du, J.~Pu, and J.~Feng, ``Ul-slam: A universal monocular line-based slam via unifying structural and non-structural constraints,'' \emph{IEEE Transactions on Automation Science and Engineering}, pp. 1--18, 2024.

\bibitem{10254455}
Z.~Zhang, Y.~Jiao, R.~Xiong, and Y.~Wang, ``Fusing multiple isolated maps to visual inertial odometry online: A consistent filter,'' \emph{IEEE Transactions on Automation Science and Engineering}, vol.~21, no.~4, pp. 5623--5638, 2024.

\bibitem{9926038}
X.~Zuo, M.~Zhang, M.~Wang, Y.~Chen, G.~Huang, Y.~Liu, and M.~Li, ``Visual-based kinematics and pose estimation for skid-steering robots,'' \emph{IEEE Transactions on Automation Science and Engineering}, vol.~21, no.~1, pp. 91--105, 2024.

\bibitem{zhangLOAMLidarOdometry2014}
J.~Zhang and S.~Singh, ``{{LOAM}}: {{Lidar Odometry}} and {{Mapping}} in {{Real-time}},'' in \emph{Robotics: {{Science}} and {{Systems X}}}.\hskip 1em plus 0.5em minus 0.4em\relax {Robotics: Science and Systems Foundation}, Jul. 2014.

\bibitem{xuFASTLIOFastRobust2021}
W.~Xu and F.~Zhang, ``{{FAST-LIO}}: {{A Fast}}, {{Robust LiDAR-Inertial Odometry Package}} by {{Tightly-Coupled Iterated Kalman Filter}},'' \emph{IEEE Robotics and Automation Letters}, vol.~6, no.~2, pp. 3317--3324, Apr. 2021.

\bibitem{dellenbachCTICPRealtimeElastic2022}
P.~Dellenbach, J.-E. Deschaud, B.~Jacquet, and F.~Goulette, ``{{CT-ICP}}: {{Real-time Elastic LiDAR Odometry}} with {{Loop Closure}},'' in \emph{2022 {{International Conference}} on {{Robotics}} and {{Automation}} ({{ICRA}})}, May 2022, pp. 5580--5586.

\bibitem{9765591}
G.~P.~C. Júnior, A.~M.~C. Rezende, V.~R.~F. Miranda, R.~Fernandes, H.~Azpúrua, A.~A. Neto, G.~Pessin, and G.~M. Freitas, ``Ekf-loam: An adaptive fusion of lidar slam with wheel odometry and inertial data for confined spaces with few geometric features,'' \emph{IEEE Transactions on Automation Science and Engineering}, vol.~19, no.~3, pp. 1458--1471, 2022.

\bibitem{10721206}
C.~Xia, X.~Li, F.~He, S.~Li, and Y.~Zhou, ``Accurate and rapidly-convergent gnss/ins/lidar tightly-coupled integration via invariant ekf based on two-frame group,'' \emph{IEEE Transactions on Automation Science and Engineering}, pp. 1--14, 2024.

\bibitem{abu-alrubRadarOdometryAutonomous2023}
N.~J. {Abu-Alrub} and N.~A. Rawashdeh, ``Radar {{Odometry}} for {{Autonomous Ground Vehicles}}: {{A Survey}} of {{Methods}} and {{Datasets}},'' Jul. 2023.

\bibitem{adolfssonCFEARRadarodometryConservative2021}
D.~Adolfsson, M.~Magnusson, A.~Alhashimi, A.~J. Lilienthal, and H.~Andreasson, ``{{CFEAR Radarodometry}} - {{Conservative Filtering}} for {{Efficient}} and {{Accurate Radar Odometry}},'' in \emph{2021 {{IEEE}}/{{RSJ International Conference}} on {{Intelligent Robots}} and {{Systems}} ({{IROS}})}, Sep. 2021, pp. 5462--5469.

\bibitem{adolfssonLidarLevelLocalizationRadar2022}
------, ``Lidar-{{Level Localization With Radar}}? {{The CFEAR Approach}} to {{Accurate}}, {{Fast}}, and {{Robust Large-Scale Radar Odometry}} in {{Diverse Environments}},'' \emph{IEEE Transactions on Robotics}, pp. 1--20, 2022.

\bibitem{zhangScanDenoisingNormal2023a}
R.~Zhang, Y.~Zhang, D.~Fu, and K.~Liu, ``Scan {{Denoising}} and {{Normal Distribution Transform}} for {{Accurate Radar Odometry}} and {{Positioning}},'' \emph{IEEE Robotics and Automation Letters}, vol.~8, no.~3, pp. 1199--1206, Mar. 2023.

\bibitem{lubancoR3RobustRadon2024}
D.~L.~S. Lubanco, A.~Hashem, M.~{Pichler-Scheder}, A.~Stelzer, R.~Feger, and T.~Schlechter, ``R{\textsuperscript{3}} {{O}}: {{Robust Radon Radar Odometry}},'' \emph{IEEE Transactions on Intelligent Vehicles}, vol.~9, no.~1, pp. 231--246, Jan. 2024.

\bibitem{limORORAOutlierRobustRadar2023a}
H.~Lim, K.~Han, G.~Shin, G.~Kim, S.~Hong, and H.~Myung, ``{{ORORA}}: {{Outlier-Robust Radar Odometry}},'' in \emph{2023 {{IEEE International Conference}} on {{Robotics}} and {{Automation}} ({{ICRA}})}, May 2023, pp. 2046--2053.

\bibitem{zhuang4DIRIOM4D2023a}
Y.~Zhuang, B.~Wang, J.~Huai, and M.~Li, ``{{4D iRIOM}}: {{4D Imaging Radar Inertial Odometry}} and {{Mapping}},'' \emph{IEEE Robotics and Automation Letters}, vol.~8, no.~6, pp. 3246--3253, Jun. 2023.

\bibitem{zhang4DRadarSLAM4DImaging2023c}
J.~Zhang, H.~Zhuge, Z.~Wu, G.~Peng, M.~Wen, Y.~Liu, and D.~Wang, ``{{4DRadarSLAM}}: {{A 4D Imaging Radar SLAM System}} for {{Large-scale Environments}} based on {{Pose Graph Optimization}},'' in \emph{2023 {{IEEE International Conference}} on {{Robotics}} and {{Automation}} ({{ICRA}})}, May 2023, pp. 8333--8340.

\bibitem{luEfficientDeepLearning4D2024}
S.~Lu, G.~Zhuo, L.~Xiong, X.~Zhu, L.~Zheng, Z.~He, M.~Zhou, X.~Lu, and J.~Bai, ``Efficient {{Deep-Learning 4D Automotive Radar Odometry Method}},'' \emph{IEEE Transactions on Intelligent Vehicles}, vol.~9, no.~1, pp. 879--892, Jan. 2024.

\bibitem{zhuo4DRVONetDeep4D2024}
G.~Zhuo, S.~Lu, H.~Zhou, L.~Zheng, M.~Zhou, and L.~Xiong, ``{{4DRVO-Net}}: {{Deep 4D Radar}}--{{Visual Odometry Using Multi-Modal}} and {{Multi-Scale Adaptive Fusion}},'' \emph{IEEE Transactions on Intelligent Vehicles}, vol.~9, no.~6, pp. 5065--5079, Jun. 2024.

\bibitem{parkPhaRaODirectRadar2020}
Y.~S. Park, Y.-S. Shin, and A.~Kim, ``{{PhaRaO}}: {{Direct Radar Odometry}} using {{Phase Correlation}},'' in \emph{2020 {{IEEE International Conference}} on {{Robotics}} and {{Automation}} ({{ICRA}})}.\hskip 1em plus 0.5em minus 0.4em\relax Paris, France: IEEE, May 2020, pp. 2617--2623.

\bibitem{barnesMaskingMovingLearning2020b}
D.~Barnes, R.~Weston, and I.~Posner, ``Masking by {{Moving}}: {{Learning Distraction-Free Radar Odometry}} from {{Pose Information}},'' in \emph{Proceedings of the {{Conference}} on {{Robot Learning}}}.\hskip 1em plus 0.5em minus 0.4em\relax PMLR, May 2020, pp. 303--316.

\bibitem{westonFastMbyMLeveragingTranslational2022a}
R.~Weston, M.~Gadd, D.~De~Martini, P.~Newman, and I.~Posner, ``Fast-{{MbyM}}: {{Leveraging Translational Invariance}} of the {{Fourier Transform}} for {{Efficient}} and {{Accurate Radar Odometry}},'' in \emph{2022 {{International Conference}} on {{Robotics}} and {{Automation}} ({{ICRA}})}, May 2022, pp. 2186--2192.

\bibitem{kungNormalDistributionTransformBased2021}
P.-C. Kung, C.-C. Wang, and W.-C. Lin, ``A {{Normal Distribution Transform-Based Radar Odometry Designed For Scanning}} and {{Automotive Radars}},'' in \emph{2021 {{IEEE International Conference}} on {{Robotics}} and {{Automation}} ({{ICRA}})}.\hskip 1em plus 0.5em minus 0.4em\relax Xi'an, China: IEEE, May 2021, pp. 14\,417--14\,423.

\bibitem{alhashimiBFARImprovingRadar2024}
A.~Alhashimi, D.~Adolfsson, H.~Andreasson, A.~Lilienthal, and M.~Magnusson, ``{{BFAR}}: Improving radar odometry estimation using a bounded false alarm rate detector,'' \emph{Autonomous Robots}, vol.~48, no.~8, p.~29, Dec. 2024.

\bibitem{hongRadarSLAMRadarBased2020b}
Z.~Hong, Y.~Petillot, and S.~Wang, ``{{RadarSLAM}}: {{Radar}} based {{Large-Scale SLAM}} in {{All Weathers}},'' in \emph{2020 {{IEEE}}/{{RSJ International Conference}} on {{Intelligent Robots}} and {{Systems}} ({{IROS}})}.\hskip 1em plus 0.5em minus 0.4em\relax Las Vegas, NV, USA: IEEE, Oct. 2020, pp. 5164--5170.

\bibitem{hongRadarSLAMRobustSimultaneous2022}
Z.~Hong, Y.~Petillot, A.~Wallace, and S.~Wang, ``{{RadarSLAM}}: {{A}} robust simultaneous localization and mapping system for all weather conditions,'' \emph{The International Journal of Robotics Research}, vol.~41, no.~5, pp. 519--542, Apr. 2022.

\bibitem{barnesRadarLearningPredict2020}
D.~Barnes and I.~Posner, ``Under the {{Radar}}: {{Learning}} to {{Predict Robust Keypoints}} for {{Odometry Estimation}} and {{Metric Localisation}} in {{Radar}},'' in \emph{2020 {{IEEE International Conference}} on {{Robotics}} and {{Automation}} ({{ICRA}})}, May 2020, pp. 9484--9490.

\bibitem{ronnebergerUNetConvolutionalNetworks2015}
O.~Ronneberger, P.~Fischer, and T.~Brox, ``U-{{Net}}: {{Convolutional Networks}} for {{Biomedical Image Segmentation}},'' in \emph{Medical {{Image Computing}} and {{Computer-Assisted Intervention}} -- {{MICCAI}} 2015}, N.~Navab, J.~Hornegger, W.~M. Wells, and A.~F. Frangi, Eds.\hskip 1em plus 0.5em minus 0.4em\relax Cham: Springer International Publishing, 2015, vol. 9351, pp. 234--241.

\bibitem{burnettWeNeedCompensate2021}
K.~Burnett, A.~P. Schoellig, and T.~D. Barfoot, ``Do {{We Need}} to {{Compensate}} for {{Motion Distortion}} and {{Doppler Effects}} in {{Spinning Radar Navigation}}?'' \emph{IEEE Robotics and Automation Letters}, vol.~6, no.~2, pp. 771--778, Apr. 2021.

\bibitem{yangTEASERFastCertifiable2021}
H.~Yang, J.~Shi, and L.~Carlone, ``{{TEASER}}: {{Fast}} and {{Certifiable Point Cloud Registration}},'' \emph{IEEE Transactions on Robotics}, vol.~37, no.~2, pp. 314--333, Apr. 2021.

\bibitem{liangScalableFrameworkRobust2020}
Y.~Liang, S.~Muller, D.~Schwendner, D.~Rolle, D.~Ganesch, and I.~Schaffer, ``A {{Scalable Framework}} for {{Robust Vehicle State Estimation}} with a {{Fusion}} of a {{Low-Cost IMU}}, the {{GNSS}}, {{Radar}}, a {{Camera}} and {{Lidar}},'' in \emph{2020 {{IEEE}}/{{RSJ International Conference}} on {{Intelligent Robots}} and {{Systems}} ({{IROS}})}.\hskip 1em plus 0.5em minus 0.4em\relax Las Vegas, NV, USA: IEEE, Oct. 2020, pp. 1661--1668.

\bibitem{dearaujoNovelMethodLand2023}
P.~R.~M. De~Araujo, M.~Elhabiby, S.~Givigi, and A.~Noureldin, ``A {{Novel Method}} for {{Land Vehicle Positioning}}: {{Invariant Kalman Filters}} and {{Deep-Learning-Based Radar Speed Estimation}},'' \emph{IEEE Transactions on Intelligent Vehicles}, vol.~8, no.~9, pp. 4275--4286, Sep. 2023.

\bibitem{rubleeORBEfficientAlternative2011}
E.~Rublee, V.~Rabaud, K.~Konolige, and G.~Bradski, ``{{ORB}}: {{An}} efficient alternative to {{SIFT}} or {{SURF}},'' in \emph{2011 {{International Conference}} on {{Computer Vision}}}.\hskip 1em plus 0.5em minus 0.4em\relax Barcelona, Spain: IEEE, Nov. 2011, pp. 2564--2571.

\bibitem{cenPreciseEgoMotionEstimation2018}
S.~H. Cen and P.~Newman, ``Precise {{Ego-Motion Estimation}} with {{Millimeter-Wave Radar Under Diverse}} and {{Challenging Conditions}},'' in \emph{2018 {{IEEE International Conference}} on {{Robotics}} and {{Automation}} ({{ICRA}})}, May 2018, pp. 6045--6052.

\bibitem{cenRadaronlyEgomotionEstimation2019}
------, ``Radar-only ego-motion estimation in difficult settings via graph matching,'' in \emph{2019 {{International Conference}} on {{Robotics}} and {{Automation}} ({{ICRA}})}, May 2019, pp. 298--304.

\bibitem{kimMulRanMultimodalRange2020}
G.~Kim, Y.~S. Park, Y.~Cho, J.~Jeong, and A.~Kim, ``{{MulRan}}: {{Multimodal Range Dataset}} for {{Urban Place Recognition}},'' in \emph{2020 {{IEEE International Conference}} on {{Robotics}} and {{Automation}} ({{ICRA}})}, May 2020, pp. 6246--6253.

\bibitem{burnettBoreasMultiSeasonAutonomous2023}
K.~Burnett, D.~J. Yoon, Y.~Wu, A.~Z. Li, H.~Zhang, S.~Lu, J.~Qian, W.-K. Tseng, A.~Lambert, K.~Y.~K. Leung, A.~P. Schoellig, and T.~D. Barfoot, ``Boreas: {{A Multi-Season Autonomous Driving Dataset}},'' Jan. 2023.

\bibitem{geigerAreWeReady2012b}
A.~Geiger, P.~Lenz, and R.~Urtasun, ``Are we ready for autonomous driving? {{The KITTI}} vision benchmark suite,'' in \emph{2012 {{IEEE Conference}} on {{Computer Vision}} and {{Pattern Recognition}}}, Jun. 2012, pp. 3354--3361.

\bibitem{behleyEfficientSurfelBasedSLAM2018b}
J.~Behley and C.~Stachniss, ``Efficient {{Surfel-Based SLAM}} using {{3D Laser Range Data}} in {{Urban Environments}},'' in \emph{Robotics: {{Science}} and {{Systems XIV}}}.\hskip 1em plus 0.5em minus 0.4em\relax {Robotics: Science and Systems Foundation}, Jun. 2018.

\end{thebibliography}

\vfill

\end{document}